%% file: bare_jrnl_compsoc.tex
\documentclass[10pt,journal,compsoc]{IEEEtran}
\usepackage{multirow}
\usepackage{float}
\usepackage[pagebackref,breaklinks,colorlinks]{hyperref}
\hypersetup{citecolor=[RGB]{119,185,0}}
\usepackage{breakcites}
\usepackage{amssymb}
\usepackage{graphicx}
\usepackage{amsmath}
\usepackage{subfig}
\usepackage[T1]{fontenc}

\ifCLASSOPTIONcompsoc
  \usepackage[nocompress]{cite}
\else
  \usepackage{cite}
\fi
\ifCLASSINFOpdf
\else
\fi
\begin{document}
\title{
%Decision Transformer: A Literature Review
On Transforming Reinforcement Learning with Transformers:
The Development Trajectory
}

\author{Shengchao~Hu,%~\IEEEmembership{Member,~IEEE,}
        Li~Shen, %~\IEEEmembership{Fellow,~OSA,}
        Ya Zhang, 
        Yixin Chen,~\IEEEmembership{Fellow,~IEEE}, 
        ~and~Dacheng~Tao,~\IEEEmembership{Fellow,~IEEE}% <-this % stops a space

\IEEEcompsocitemizethanks{
\IEEEcompsocthanksitem This work is supported by Science and Technology Innovation 2030 –“Brain Science and Brain-like Research” key Project (No. 2021ZD0201405).
\IEEEcompsocthanksitem Shengchao Hu is with Shanghai Jiaotong University, China. %\protect\\
E-mail: charles-hu@sjtu.edu.cn
\IEEEcompsocthanksitem Li Shen is with JD Explore Adacemy, China. 
Email: mathshenli@gmail.com
\IEEEcompsocthanksitem Ya Zhang is with  Shanghai Jiaotong University, China. Email: ya\_zhang@sjtu.edu.cn
\IEEEcompsocthanksitem Yixin Chen is with  Washington University in St Louis, America. Email: chen@cse.wustl.edu
\IEEEcompsocthanksitem Dacheng Tao is with JD Explore Adacemy, China and the University of Sydney, Australia. 
Email: dacheng.tao@gmail.com
}% <-this % stops an unwanted space
}


\markboth{Journal of \LaTeX\ Class Files,~Vol.~14, No.~8, August~2022}%
{Shell \MakeLowercase{\textit{et al.}}: Bare Demo of IEEEtran.cls for Computer Society Journals}



\IEEEtitleabstractindextext{%
\begin{abstract}
%Transformer, originally used in the field of natural language processing, has also been widely used in the field of computer vision due to its excellent representation ability.
Transformers, originally devised for natural language processing (NLP), have also produced significant successes in computer vision (CV).
%and recently attracted serious attention in reinforcement learning (RL), all due to its excellent representation ability. 
Due to their strong expression power, researchers are investigating ways to deploy transformers for reinforcement learning (RL), and transformer-based models have manifested their potential in representative RL benchmarks.
%
%With the rapid adoption of deep reinforcement learning (RL), especially the success of offline RL, researchers are looking for ways to apply transformers to the RL field, leveraging its powerful representation capabilities to improve overall performance.
%In a variety of RL benchmarks, transformer-based decision models perform similar to or better than traditional deep RL algorithms,
%
%Though it is promising and attracts many researchers' attention, there is currently a lack of summary/guidelines for the development of decision transformers.
%
%Initially, transformer is usually used as an alternative architecture and mainly used to provide the memory information,
%\ls{its connection with the following sentence should be strength.}. 
%
%while treating the decision making progress as the sequence modeling with transformer demonstrates impressive performance and receives more and more attention from the RL community.
%
In this paper, we collect and dissect recent advances concerning the transformation of RL with transformers (transformer-based RL (TRL)) to explore the development trajectory and future trends of this field.
%
%Specifically, we classify existing decision transformer methods as two categories: trajectory optimization and architecture enhancement, and summarize the main applications into robotic manipulation, text-based games, navigation and autonomous driving.  
%considering that there are currently two main types of methods using decision transformers, 
We group the existing developments into two categories: architecture enhancements and trajectory optimizations, and examine the main applications of TRL in robotic manipulation, text-based games (TBGs), navigation and autonomous driving.
Architecture enhancement methods consider how to apply the powerful transformer structure to RL problems under the traditional RL framework, and they model agents and environments much more precisely than deep RL methods. However, these methods still limited by the inherent defects of traditional RL algorithms, such as bootstrapping and the “deadly triad”.
Trajectory optimization methods treat RL problems as sequence modeling problems and train a joint state-action model over entire trajectories under the behavior cloning framework; such approaches are able to extract policies from static datasets and fully use the long-sequence modeling capabilities of transformers.
%but sometimes it would trap in sub-optimum due to the strict conditions.
%introduce how to leverage the transformer architecture to solve RL problems via the perspective of sequence modeling.
%\ls{sequence modeling has been used before. Replace it with another words. Here, we should interpret "sequence modeling category"}
%
%focus on the transformer usage in traditional RL algorithms \ls{explain the sequence making category}. %and further divide it into feature representation and dynamics and reward by the specific part of transformer participation.
%
%At last, we discuss the challenges and propose several future research directions for this rapidly growing field. 
Given these advancements, the extensions and challenges in TRL are reviewed and proposals regarding future research directions are discussed.
We hope that this survey can provide a detailed introduction to TRL and motivate future research in this rapidly developing field.
\end{abstract}

}

\maketitle

\IEEEdisplaynontitleabstractindextext

% \fi
\IEEEpeerreviewmaketitle

\input{texfile/1.introduction}

\input{texfile/2.preliminary}
\input{texfile/3.sequencemaking}
\input{texfile/4.sequencemodeling}
\input{texfile/5.application}
\input{texfile/6.Discussion}

\ifCLASSOPTIONcaptionsoff
  \newpage
\fi



\bibliographystyle{IEEEtran}
\bibliography{ref}
\end{document}

%% file: texfile/1.introduction.tex
\IEEEraisesectionheading{\section{Introduction}\label{sec:introduction}}
\IEEEPARstart{R}{ecently}, the transformer architecture has made substantial progress in natural language processing (NLP) tasks \cite{NLP}.
For example, generative pretraining (GPT) series models \cite{brown2020language} and bidirectional encoder representations from transformers (BERT) models \cite{bert} have achieved state-of-the-art performance on a wide range of downstream tasks (e.g. question answering (QA) and sentence classification).
%\ls{it should be slightly clear. strong performance and downstream tasks.}
%
%Though transformer is most used in NLP due to the naturally sequential property of language, it can  
Inspired by the success of the transformer architecture in NLP, researchers have also tried to apply transformers to computer vision (CV) tasks.
%
%Recently, transformer architectures further promote the development of computer vision applications.  
%
Chen et al. \cite{chen2020generative} utilized a transformer to autoregressively predict pixels, and a vision transformer (ViT) \cite{dosovitskiy2020image} was directly applied to sequences of image patches to classify full images. This approach has achieved state-of-the-art performance on multiple image recognition benchmarks.
Considering an image as a sequence of subimages \cite{dosovitskiy2020image, liu2021swin}, the transformer architecture is able to extract representative features via its self-attention mechanism \cite{bahdanau2014neural, parikh2016decomposable} and can be further used to generate more powerful multimodal visual-language models, such as DALL-E \cite{DALLE}, Flamingo \cite{flamingo}, and Gato \cite{Gato}.

\begin{figure*}[!t]
    \centering
    \includegraphics[width=7in]{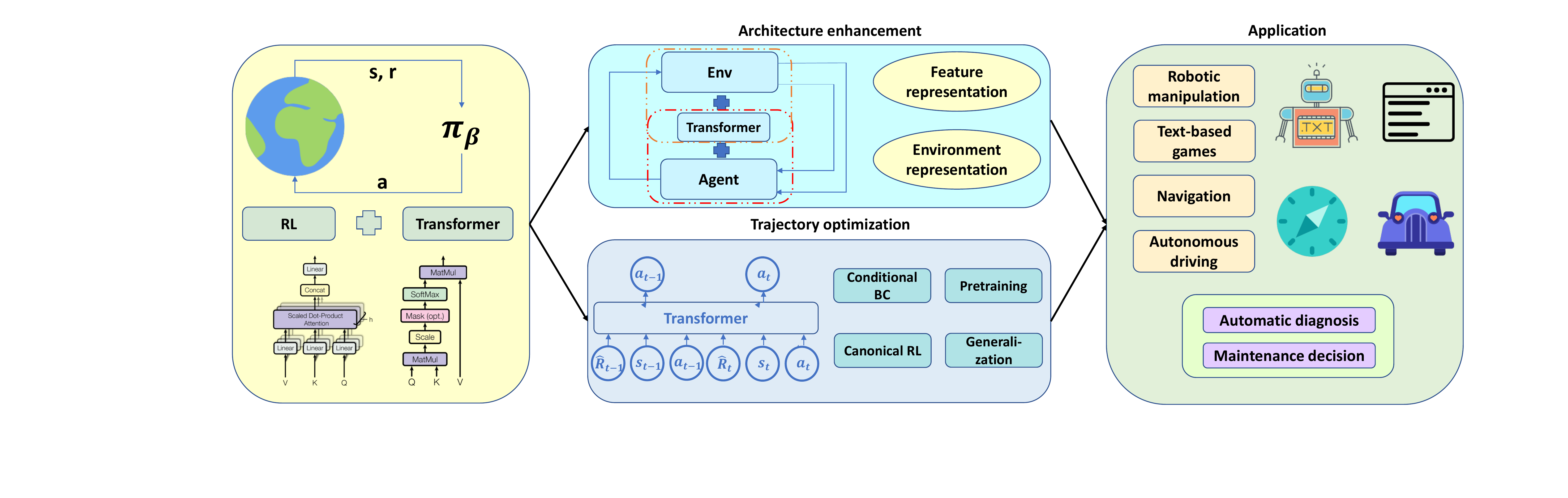}
    \caption{A detailed overview of the transformers' involvement in RL and realistic application. 
    %
    %An illustration of how decision transformer is involved in RL and realistic application is discussed in this paper. 
    The architecture enhancement block demonstrates the specific part of transformer participation in traditional RL methods.
    The trajectory optimization block shows the usage of the transformer and corresponding research categories.
    Finally, the application block introduces several realistic applications of TRL. (e.g. robotic manipulation, text-based games, and so on.)}
    \label{fig:pipeline}
\end{figure*}

Reinforcement learning \cite{sutton2018rlintroduction} (RL) is a powerful control strategy that usually consists of an environment and an agent.
The agent observes the current state of the environment, makes actions, and obtains the reward for the current action and the state of the next moment; the goal of the agent is to maximize the cumulative reward it obtains.
With the success of deep neural networks, deep RL, which can make use of the high-capacities capabilities of deep neural networks to make function approximation methods more accurate and enable agents to operate normally in unstructured environments, has gradually become popular.
%
%While the previously dominant networks used in deep RL were mainly simple MLP and CNN networks \cite{DQN, SAC} due to the simple test environment and limited training data\ls{the connection with these two sentences should be strength.}.
%
%On the one hand, because the test environment itself is not particularly complicated and usually only considering a single task, the simple network is enough for the agent to remember all relevant information. 
%
%On the other hand, the previous algorithms need to collect data from the interaction with the environment, which limits the amount of training datasets and thus small networks is enough to converge to optimal point.
%
%A large part of the success of deep learning is attributed to large and diverse datasets \ls{here, we are talking about deep RL. We should mention deep RL first.}, and the current deep RL mainly relies on interacting with the environment to dynamically collect data, which largely hinders the application of deep RL in real world.
%
%
However, current deep RL approaches mainly rely on interacting with the environment to dynamically collect data, which limits the amount of training data collected and may result in difficulty when it is expensive to interact with the environment.
This occurs in cases such as robotic manipulation \cite{singh2022reinforcement}, education \cite{singla2021reinforcement}, healthcare \cite{liu2020reinforcement}, and autonomous driving \cite{kiran2021deep}. 
Thus, offline RL \cite{levine2020offline, prudencio2022survey} has attracted researchers' interest as a data-driven learning method that can extract policies from static previously collected datasets without interacting with the environment.
%
%This approach is extremely valuable in conditions where it is very difficult or expensive to interact with the environment, such as the robotic manipulation \cite{singh2022reinforcement}, education \cite{singla2021reinforcement}, healthcare \cite{liu2020reinforcement} and autonomous driving \cite{kiran2021deep}. 
%
The offline RL setting can make full use of the ability of deep networks to extract the optimal policy from a large amount of offline data, but it also needs to address the distribution discrepancy between the offline training data and the target policy.

%
% After learning the offline policy, one can further optimize it in the online environment, and use the offline policy as the initial strategy for the subsequent learning process.
%
% Compared with random initialization, it usually has higher sampling efficiency and security.
% \ls{too many background for RL}

Due to the sequential decision process of RL, a natural idea is to apply transformers, which have been fully developed in recent years \cite{GTrXL, siebenborn2022crucial}, to augment deep RL methods.
Initially, transformers were typically used as alternative architectures for replacing convolutional neural networks (CNNs) or long short-term memory (LSTM) in deep RL methods, and they mainly provided memory information for the agent network in environments where the critical observations often spanned the entire episode.
However, these methods suffer from the instability issue from an empirical standpoint when directly applying the original transformer structure to the decision process \cite{mishra2017simple}; thus, GtrXL \cite{GTrXL} first focused on modifying the transformer architecture to better adapt it to RL, and it has motivated follow-up works that have provided more powerful transformer architectures \cite{rfwps, rae2019compressive}.
The decision transformer (DT) \cite{DT} and trajectory transformer (TT) \cite{TT} simultaneously treat the RL problem as a conditional sequence modeling problem and leverage the transformer architecture to model sequential trajectories, taking advantage of the powerful long-sequence modeling capability of the transformer architecture and inspiring many subsequent works.
%
%Following these pioneering works, a series of new studies have been proposed to solve different RL tasks with different techniques.
%
However, the current works enhance the performance of transformer-based RL (TRL) methods from various perspectives, and there is a lack of a unified understanding and summary of TRL.
To keep pace with the rate of new progress and promote the development of TRL, a survey of existing works would be helpful for researchers in the community.

In this paper, we focus on providing a comprehensive overview of recent advances in TRL and discussing the challenges and potential directions for future improvement (shown in Figure \ref{fig:pipeline}).
To facilitate future research on different topics, we categorize the developed TRL approaches from the perspective of their methods and applications, as listed in Table \ref{Taxonomy}, which include \textbf{architecture enhancements}, \textbf{trajectory optimizations}, and \textbf{applications}.

Regarding architecture enhancement, we mainly consider how to apply a more powerful transformer structure to RL problems under the traditional RL framework.
According to the specific part of the process in which transformers participate, we further divide these methods into feature representation and environmental representation techniques.
Feature representation approaches mainly includes methods where a transformer is used to extract feature representations from a multimodal input, and then an agent utilizes these representations to make decisions through value- or policy-based methods.
Environmental representation mainly illustrates how to leverage the transformer architecture for dynamics and reward modeling, which can be used to form better decision processes with additional planning algorithms.

Trajectory optimization methods all regard RL as a sequence modeling problem and use decision transformers to learn policies under the behavior cloning framework.
According to their motivations and target tasks, we further classify trajectory optimization methods into conditional BC, canonical RL, pretraining, and generalization approaches.
1) Conditional BC treats RL as conditional behavior cloning and adopts the transformer architecture to model state-action-reward sequence; these techniques include the pioneering works and their conditions for optimality.
% and separates the pioneering works from other follow-on works.
% %
2) Canonical RL attempts to combine the advantages of sequence modeling methods and traditional RL algorithms, significantly improving the performance of the original algorithms.
% %
3) Pretraining focuses on how to pretrain decision transformers so that they can achieve better performance in downstream RL tasks, especially in terms of their pretraining datasets and objectives.
%Since the pretraining technique has been widely studied in the NLP field, there are many works to study whether the pretraining technique can continue to play a role in the RL field, which are classified into category pretraining.\ls{we should directly show what is the pretrain in decision transformer and how does it work... }
%
%Category pretraining emphasizes the role of pretraining in the decision transformer, and points out that pre-training on data in other domains (e.g. NLP) can also have a positive effect.
% %
%Category generalization deals with the generalization ability of decision transformer, especially for the multi-task setting and transfer setting.
4) Generalization mainly involves algorithms for which a single agent is able to perform well in multiple environments, which may include unseen environments. 
%how to achieve generalization in the RL field through the method of decision transformer, especially for the multi-task setting and transfer learning setting.
% %
%5) Limitation gives several limitations of the decision transformers and points out the conditions for DT to converge to the optimal point. 
%
%
% While for the architecture enhancement, we mainly consider how to apply a more powerful transformer structure to RL problems under the traditional RL framework.
% %
% According to the specific part of transformer participation, we further divide it into feature representation and environment representation.
% %
% Feature representation mainly introduces the methods where the transformer is used to extract feature representations from the multi-modal input and then the agent utilizes these representations to make decisions through value or policy based methods.
% %
% Environment representation mainly illustrates how to leverage the transformer architecture for the dynamics and reward modeling, which can be used for better decision process with additional planning algorithms.
%
Finally, we summarize four types of typical applications of TRL, including robotic manipulation, text-based games (TBGs), vision-language navigation, and autonomous driving, each of which is highly related to TRL and can be solved by RL methods.
Toward the end of this paper, we introduce TRL to the multiagent RL (MARL) extensions, discuss several challenges and provide insights into the future prospects of this line of research.

The rest of the paper is organized as follows.
Section \ref{sec:preliminaries} introduces the preliminaries of this survey, including a brief overview of RL and the foundation of the standard transformer.
Sections \ref{sec:DT} and \ref{sec:application} are the main parts of the paper, in which we summarize the existing TRL methods and their applications, respectively.
Then, in Section \ref{sec:ExtensionDiscussion}, we briefly describe TRL in the MARL domain, which is an extension of single-agent RL, and discuss several future directions and challenges related to TRL.
%
%Finally we give our conclusions and discuss several research directions and challenges based on these summarized researches.
Finally, we provide a summary to conclude this work.
Note that in this survey, we mainly include the representative works, and some recent preprint works on arXiv may be missed due to the rapid development of this field.

% intro的第一段：讲 transformer 在 nlp和cv里面的丰功伟绩
% intro的第二段：讲 transformer 开始在 RL上展露头脚，目前出现了大量的论文从不同的角度把 transformer应用到RL中，但是现在大家的方式都很杂，对这个领域缺乏一个统一的理解和总结。（motivation）
% intro的第三段：在本文中，我们总结现有的 decision transformer的工作，根据XXX将其分为多少个类别。 然后分别介绍各个类别表示的是什么含义（仔细说清楚我们是怎么分类的）； 另外
% intro的第四段：根据目前的总结，我们额外给出了一些潜在的研究方向和挑战。然后大概说一下
% intro的第五段： 最后说一下本文的organization

\begin{figure*}[!t]
    \centering
    \includegraphics[width=7in]{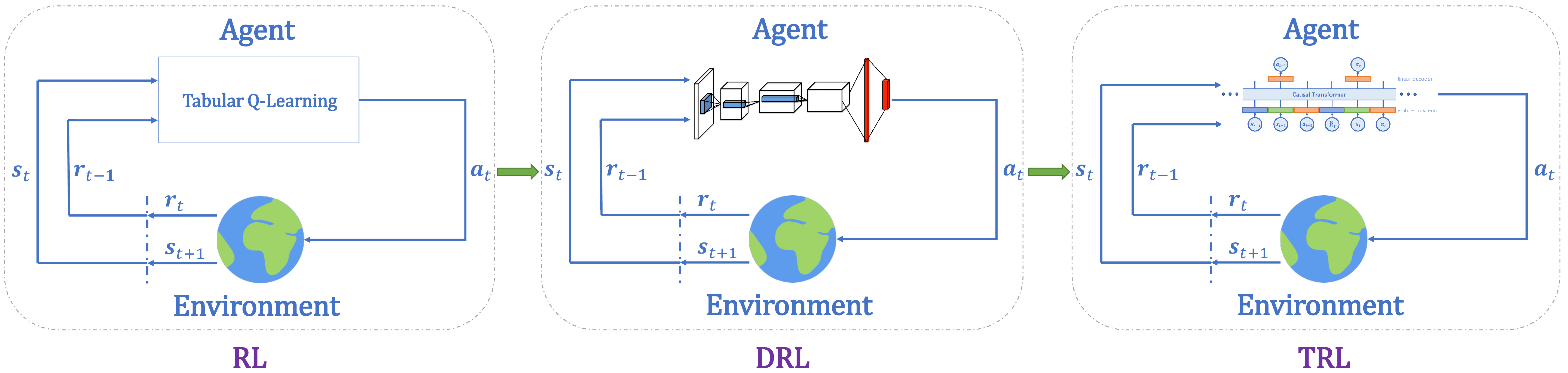}
    \caption{The development trajectory of RL. 
    At first, the RL agents use the tabular Q-learning to make decisions at the current moment, then the DRL agents begin to use the deep network to estimate the value and policy functions, and finally, the TRL agents leverage the ability of transformer architecture to evaluate the policy.
    There is a similar trend for environment modeling in the model-based RL.}
    \label{fig:d_trajectory}
\end{figure*}

\begin{table*}[]
\caption{Representative works of TRL}
\label{Taxonomy}
\centering
\scalebox{0.9}{
\begin{tabular}{c|c|ccc}
\hline
\textbf{Category} &
  \textbf{Sub-category} &
  \textbf{Method} &
  \textbf{Highlights} &
  \textbf{Publication} \\ \hline
\multirow{10}{*}{\begin{tabular}[c]{@{}c@{}}Architecture \\ Enhancement\end{tabular}} &
  \multirow{5}{*}{Feature Representation} &
  GTrXL \cite{GTrXL} &
  Replace LSTM in V-MPO with Gated Transformer-XL &
  ICML 2020 \\
 &
   &
  WMG \cite{WMG}&
  Transformer-based RL with factored observations &
  ICML 2020 \\
 &
   &
  RFWPs \cite{rfwps} &
  Add recurrence to the FWP &
  NeurIPS 2021 \\
 &
   &
  ALD \cite{ALD} &
  Actor-Learner Distillation &
  ICLR 2021 \\
 &
   &
  COBERL \cite{coberl} &
  Combine contrastive loss and a hybird LSTM-transformer &
  ICLR 2022 \\ \cline{2-5} 
 &
  \multirow{5}{*}{Environment Representation} &
  IRIS \cite{IRIS} &
  World model composed of a autoencoder and a transformer &
  arXiv 2022 \\
 &
   &
  TransDreamer \cite{transdreamer} &
  Transformer-based stochastic world model &
  arXiv 2022 \\
 &
   &
  VQM \cite{vqm} &
  Latent representations transition model &
  ICML 2021 \\
 &
   &
  PlaTe \cite{plate} &
  Learn state and action spaces from videos &
  arXiv 2021 \\
 &
   &
  MINECLIP \cite{minedojo} &
  An automatic evaluation metric &
  arXiv 2022 \\ \hline
\multirow{21}{*}{\begin{tabular}[c]{@{}c@{}}Trajectory \\ Optimization\end{tabular}} &
  \multirow{4}{*}{Conditioned BC} &
  DT\cite{DT} &
  Return-to-go, BC conditioning desired RTG &
  NeurIPS 2021 \\
 &
   &
  TT\cite{TT} &
  Modeling distribution of trajectory, beam search &
  NeurIPS 2021 \\ 
 &
   &
  ESPER \cite{esper} &
  Cluster trajectories, average cluster returns &
  arXiv 2022 \\
 &
   &
  RCSL \cite{rcsl} &
  Theoretical analysis for capabilities and limitations &
  NeurIPS 2022 \\ \cline{2-5} 
 &
  \multirow{5}{*}{Canonical RL} &
  ODT\cite{odt} &
  Blends offline pretraining with online finetuning &
  ICML 2022 \\
 &
   &
  StARformer\cite{starformer} &
  Markovian-like inductive bias &
  ECCV 2022 \\
 &
   &
  AD\cite{AD} &
  DT with in-context reinforcement learning &
  arXiv 2022 \\
 &
   &
  BooT\cite{BooT} &
  Self-generate data &
  NeruIPS 2022 \\
 &
   &
  QDT\cite{QDT} &
  Utilize Q function to relabel the RTG &
  arXiv 2022 \\ \cline{2-5} 
 &
  \multirow{5}{*}{Pretraining} &
  ChibiT\cite{chibiT} &
  Pre-train on the Wikipedia &
  arXiv 2022 \\
 &
   &
  LID\cite{LID} &
  Pre-trained LMs as a general scaffold &
  NeruIPS 2022 \\
 &
   &
  Effect of Pre-training\cite{Takagi2022on} &
  Investigate the effect of pre-training &
  NeruIPS 2022 \\
 &
   &
  MaskDP\cite{MaskDP} &
  A pretraining method to learn generalizable models &
  NeurIPS 2022 \\
 &
   &
  ACL\cite{ACL} &
  What representation matters in the offline pretraining &
  ICML 2021 \\ \cline{2-5} 
 &
  \multirow{7}{*}{Generalization} &
  Multi-Game DT \cite{multigame_DT} &
  Power-law performance trend &
  NeuIPS 2022 \\
 &
   &
  Gato\cite{Gato} &
  Multi-modal, multi-task, multi-embodiment model&
  arXiv 2022 \\
 &
   &
  Switch TT\cite{switchtt} &
  Sparsely activated model and trajectory value estimator &
  arXiv 2022 \\
 &
   &
  GDT \cite{GDT} &
  Hindsight information matching generalization&
  ICLR 2022 \\
 &
   &
  FlexiBiT \cite{unimask} &
  Randomly masking and predicting &
  NeurIPS 2022 \\
 &
   &
  Prompt-DT \cite{Prompt-DT} &
  Trajectory prompt for generalization&
  ICML 2022 \\
 &
   &
  Transfer-DT \cite{transfer-DT} &
  Changes in the environment dynamics, causal reasoning &
  arXiv 2021 \\ \hline
\multirow{17}{*}{Application} &
  \multirow{5}{*}{Robotic Manipulation} &
  T-OSIL \cite{T-OSIL}  &
  Transformer for OSIL &
  CoRL 2021 \\
 &
   &
  LATTE \cite{LaTTe}&
  Modify the trajectory based on the multi-modal transformer &
  arXiv 2022 \\
 &
   &
  Hiveformer \cite{Hiveformer} &
  History-aware instruction-conditioned multi-view transformer &
  CoRL 2022 \\
 &
   &
  PERACT \cite{PERACT}&
  Language goals with voxel observation &
  CoRL 2022 \\
 &
   &
  TTP \cite{TTP} &
  Prompt-situation Transformer, multi-preference learning &
  arXiv 2022 \\ \cline{2-5} 
 &
  \multirow{4}{*}{Text-based Games} &
  GATA \cite{gata} &
  Learn to construct and update graph-structured beliefs &
  NeurIPS 2020 \\
 &
   &
  Q*BERT \cite{QBERT} &
  Build a knowledge graph by answering questions &
  arXiv 2020 \\
 &
   &
  OOTD \cite{OOTD} &
  Model-based methods for TBGs &
  ICLR 2022 \\
 &
   &
  LTL-GATA \cite{LTL-GATA} &
  Equip GATA with instruction-following capabilities &
  NeurIPS 2022 \\ \cline{2-5} 
 &
  \multirow{4}{*}{Navigation} &
  PREVALENT \cite{prevalent} &
  Pretraining with finetuning for VLN tasks &
  CVPR 2021 \\
 &
   &
  VLNBERT \cite{VLN-BERT}&
  Equip BERT model with recurrent function &
  CVPR 2021 \\
 &
   &
  HAMT \cite{HAMT} &
  Hierarchical transformer encoding long-range history &
  NeurIPS 2021 \\
 &
   &
  AVLEN \cite{avlen}&
  Agent for audio-visual-language embodied navigation &
  NeurIPS 2022 \\ \cline{2-5} 
 &
  \multirow{4}{*}{Autonomous Driving} &
  SPLT \cite{SPLT} &
  Disentangling the policy and world models &
  ICML 2022 \\
 &
   &
  BeT \cite{BeT} &
  Modeling unlabeled demonstration data with multiple modes &
  NeurIPS 2022 \\
 &
   &
  TransFuser \cite{transfuser} &
  Fusion of intermediate features of the front view and LiDAR &
  CVPR 2021 \\
 &
   &
  InterFuse \cite{InterFuser} &
  Safety-enhanced framework with multi-modal, view sensors &
  CoRL 2022 \\ \hline
\end{tabular}
}
\end{table*}

%% file: texfile/2.preliminary.tex
\section{Preliminaries}
\label{sec:preliminaries}

In this section, we review the basic concepts of RL and transformers, which form the preliminary knowledge of this survey.

\subsection{RL}

RL involves learning to select the optimal action based on the current state at every timestep to maximize the numerical reward signal earned when each state-action pair is assigned a reward according to the problem definition.
The essence of RL is to learn through environmental interaction. 
An agent interacts with its environment and receives the corresponding reward; then, the agent can alter its behavior to produce a better policy based on the interaction consequence. 
Therefore, the agent is not guided by the "correct" actions (as in supervision learning) in most settings but instead by the actions' estimated future cumulative reward since an action may affect all subsequent rewards.
What makes RL different and difficult is its trial-and-error search and delayed reward characteristics\cite{sutton2018rlintroduction}.

Below, we first illustrate the basic model formulation of a Markov decision process (MDP) and point out its essential properties.
Then, we detail the elements of RL, presenting the most commonly used components found in a typical RL algorithm. 
%
%Finally, we discuss the challenges faced in RL to describe what the latest RL research focuses on.

\subsubsection{MDPs}
As a mathematical model of a sequential decision problem, an MDP is considered the ideal modeling approach in dynamic environments and provides the theoretical basis for RL.
%
%\ls{the figure of MDP and use it to introduce action, state, and reward}
%
To be specific, as shown in Figure \ref{fig:d_trajectory}, an agent is asked to make a decision $a_t \in \mathcal{A}$ based on the current state $s_t \in \mathcal{S}$; then, the environment responds to the action made by the agent and transforms the state to the next state $s_{t+1} \in \mathcal{S}$ with the reward $r_t \in \mathbb{R}$.
% \begin{figure}[!t]
%     \centering
%     \includegraphics[width=2.5in]{figs/MDP.pdf}
%     \caption{Diagram of MDP. Observing the current state $s_t$, the agent would make decision $a_t$, then the environment would transform to the next state $s_{t+1}$ with the reward $r_t$.}
%     \label{fig:MDP}
% \end{figure}

Formally, an MDP can be defined by a 6-tuple notation $\mathcal{M} = (\mathcal{S}, \mathcal{A}, \mathcal{T}, \rho_0, \mathcal{R}, \gamma)$, where $\mathcal{S}$ is the state space, $\mathcal{A}$ is the action space, $T(s_{t+1} | s_t, a_t)$ denotes the transition probability from state $s_t$ to $s_{t+1}$ after taking action $a_t$, $\rho_0$ denotes the initial state distribution, $R(s_t,a_t)$ denotes the reward value after taking action $a_t$ at state $s_t$ and $\gamma \in (0, 1]$ denotes the discount factor.
Under the MDP setting, a trajectory is defined as the experience of the agent, which is denoted by $\tau = (s_1, a_1, r_1, s_2, a_2, r_2, \dots, s_t, a_t, r_t)$, and a policy is defined as a probability function $\pi(a_t | s_t)$, which represents the probability of taking action $a_t$ at state $s_t$.
We aim to find the optimal policy $\pi^*(a | s)$ that maximizes the expected total reward for all the possible trajectories induced by the policy such that:
\begin{equation}
    \pi^* = \arg\max_{\pi} \mathbb{E}_{\tau \sim \pi} [\sum_{t=1}^T r_t].
\end{equation}

Note that in an MDP, the environment's dynamics are described by the transition distribution $T(s_{t+1} | s_t, a_t)$, which means that the environment is only determined by the current state $s_t$ and the action $a_t$ while remaining independent of the historical information; this is called the Markov property.
Thus, the state information should capture all the necessary information for the agent to make the appropriate decision, and the policy $\pi(a_t | s_t)$ is only conditioned on the current state.

There are also some cases in which the agent cannot access the fully observable state, which are defined as partially observable MDPs (POMDPs).
In the POMDP setting, the agent only access to the local observation $o_t \in \mathcal{O}$, where the distribution of the observation $p(o_{t+1} | s_{t+1}, a_t)$ is dependent on the current state and the previous action \cite{kaelbling1998pomdp}. 
Although we can still apply the MDP solving process to the policy $\pi(a_t|o_t)$ and assume that the Markov property is also valid for these observations, a better model can be formulated as $\pi(a_t | o_t, o_{t-1}, \cdots, o_{t-h+1})$, which leverages the historical information to mitigate the impact of not being able to access the true states.
The POMDP is a common modeling method in RL tasks and real environments, such as TBGs and navigation.

\subsubsection{Elements of RL}

In addition to the agent and the environment, an RL system contains four main elements: a policy, a reward function, a value function, and a model of the environment based on the MDP definition\cite{sutton2018rlintroduction}.
The policy specifies the action of the agent given the current state, which is the core of an RL agent since it is the only knowledge source that guides the agent's action.
Whereas a reward function is an immediate return received after taking an action, a value function specifies the total reward an agent can expect to accumulate over the future given the current state.
We seek the action that induces the state with the highest value rather than the highest individual reward because we wish to obtain good final results that are correlated with the cumulative reward.
Modeling the environment with a planning method helps the agent decide its action by considering possible future situations.

Three quantities of interest are typically used to find the optimal policy through different methods: the action-value function $Q^{\pi}(s_t, a_t)$, the state-value function $V^{\pi}(s_t)$ and their difference, which is known as the advantage function $A^{\pi}(s_t, a_t)$. 
The state-value function $V^{\pi}(s_t)$ maps a state to the expected reward when starting from state $s_t$ under policy $\pi$; similarly, the action-value function $Q^{\pi}(s_t, a_t)$ maps a state-action pair to the expected reward.
The advantage function $A^{\pi}(s_t, a_t)$ is typically used to lower the variance of the action-value function to distinguish between the impacts of the state and action\cite{dueling_DQN}.

Based on these quantities, there are roughly three classes of methods: dynamic programming, model-free, and model-based methods \cite{prudencio2022survey}.
Dynamic programming is used to compute the optimal policy given a known MDP.
In model-free methods, since we do not know the MDP, we use the sampling method instead.
We can directly optimize the policy through a policy gradient method (e.g., REINFORCE \cite{REINFORCE}, PG\cite{PG}, or NPG\cite{NPG}), use a value iteration method to determine the policy from the value function (e.g., the deep Q-network (DQN) \cite{DQN}, Double DQN \cite{DDQN}, or Dueling DQN \cite{dueling_DQN}), or use an actor-critic method that adopts the critic's value to improve the actor's policy (e.g., DDPG \cite{DDPG}, PPO \cite{PPO}, or SAC\cite{SAC}).
In model-based methods (e.g., MOReL\cite{morel}, MOPO \cite{mopo}, and COMBO\cite{combo}), we learn an environment model that simulates an MDP, and then we can either use dynamic programming for planning under the simulated MDP or use model-free methods by sampling from the simulated MDP.

Recently, offline RL has attracted much interest; in this paradigm, an agent extracts return-maximizing policies from some fixed limited datasets consisting of trajectory rollouts of arbitrary policies \cite{levine2020offline}.
Formally, static datasets are defined as follows: $\mathcal{D}= \{(s_t, a_t, s_{t+1}, r_t)_i\}$, where $i$ is the index, the actions and states come from the distribution induced by the behavior policy $(s_t, a_t) \sim d^{\pi_{\beta}}(\cdot)$, and the next states and rewards are determined by the dynamics $(s_{t+1}, r_t) \sim (T(\cdot | s_t, a_t), r(s_t, a_t))$.
In contrast to most online RL methods, offline RL makes itself applicable in domains where it is currently expensive to collect data online, such as healthcare\cite{Diaformer}, inventory maintenance\cite{khorasgani2021maintenance} and autonomous driving\cite{SPLT}.
However, offline RL is harder to use since it cannot interact with the environment to explore new states and transitions; thus, RL approaches often need pessimistic values or policy constraints to restrict access to out-of-distribution phenomena.
Otherwise, when encountering unseen states, agents will make mistakes until their policy diverges greatly from the training policy, which is called a distribution shift. 

%\ls{introduce the existing popular methods for RL: DQN, PPO, A3c, SAC and related methods at the suitable position.}

% \subsubsection{Challenges in RL}
% Several basic challenges are emphasized in RL\cite{deeprl_survey}.
% %
% First, the agent needs to perform a trial-and-error search to find the optimal policy by considering the reward signal; however, this process may be expensive or encounter danger when actively interacting with a real-world environment.
% %
% Then, the state (or observation) mainly depends on the agent's action and may correlate with other agents' actions (e.g., a car driving in a lane), and the state may be correlated with the historical information, which means the environment is not strictly following the MDP formulation.
% %
% Finally, the agent needs to deal with long-term dependencies, where the consequence of an action can only be shown through many environmental transitions, which is called the (temporal) credit assignment problem \cite{sutton2018rlintroduction}.

\subsection{Transformer Architecture}
\begin{figure}[!t]
    \centering
    \includegraphics[width=2.5in]{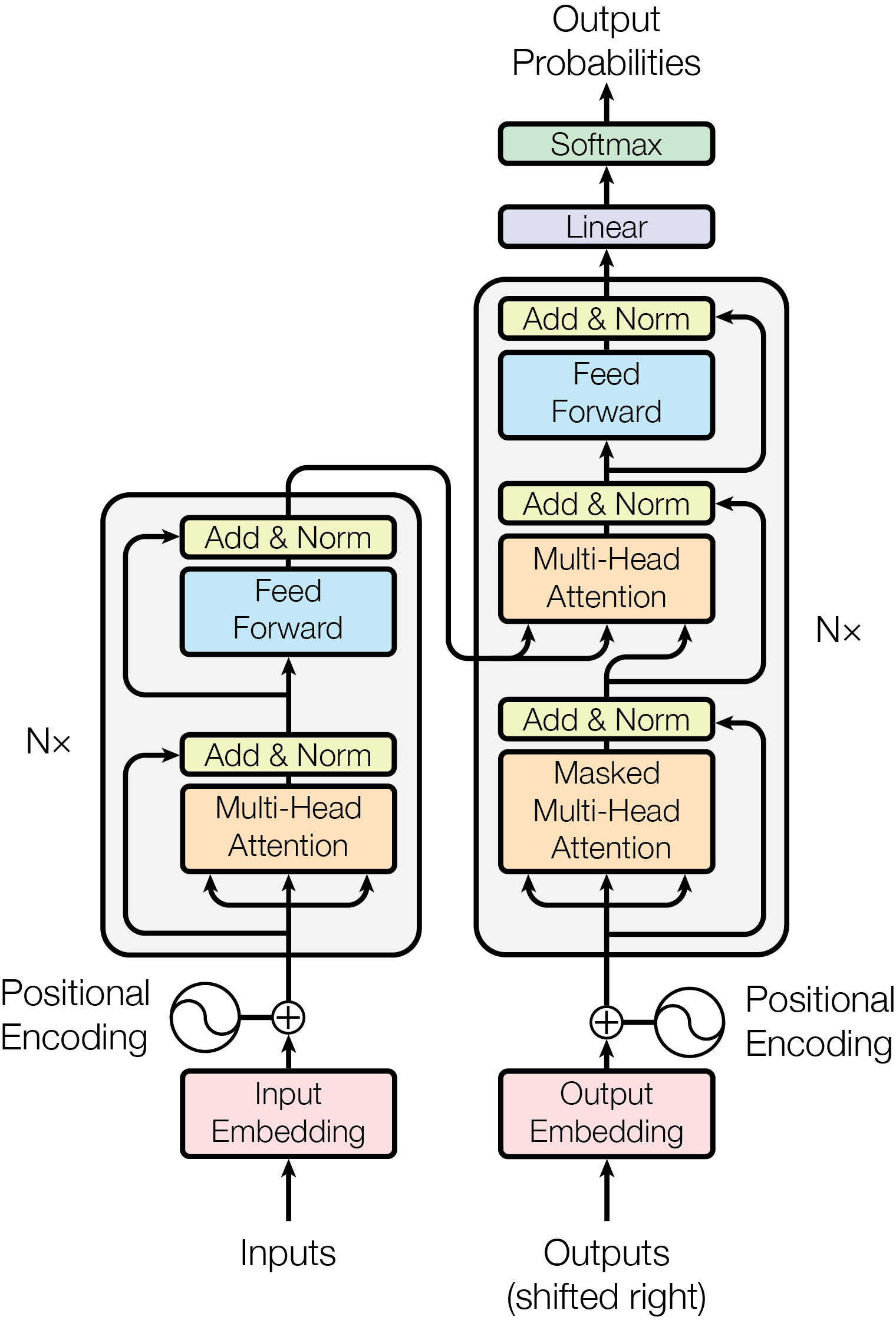}
    \caption{Structure of the original transformer (image from \cite{vaswani2017attention})}
    \label{fig:transformer}
\end{figure}

The transformer\cite{vaswani2017attention} was first proposed as a network architecture for machine learning tasks in the field of NLP.
The overall structure diagram is shown in Figure \ref{fig:transformer}; it consists of an encoder and a decoder using stacked self-attention and pointwise fully connected layers.
The encoder maps an input sequence of tokens to latent representations, and then the decoder uses this sequence to generate the desired results in an autoregressive manner.
This process consumes the previously generated results as additional inputs for generating the next output.
In the following, we describe each component of the transformer in detail.

\subsubsection{Self-Attention}
Self-attention \cite{lin2017structured,parikh2016decomposable}, which is the core component of the transformer, models the pairwise relations between tokens.
To calculate the self-attention, the input token representation is linearly mapped to three vectors: a query vector $q$, a key vector $k$, and a value vector $v$ with dimensions $d_q=d_k=d_v$.
Then, we pack the vectors derived from different inputs together into three different matrices: $Q, K, V$ and compute the attention map as follows:
\begin{equation}
    \label{eqt:attention}
    \text{Attention}(Q,K,V) = \text{softmax}(\frac{QK^T}{\sqrt{d_k}})V.
\end{equation}
Eq.\ref{eqt:attention} illustrates that we first compute the score between each pair of different input vectors and then translate these scores into probabilities via the softmax operation.
Finally, we obtain the weighted sum of the values where the weights are the corresponding probabilities.

The self-attention layers in the encoder and decoder modules are similar except that the layer in the decoder can only access positions that are less than the current predicted position, which is achieved by the masking technique.
The encoder-decoder attention layer in the decoder module allows every position in the decoder to query all positions in the input sequence, which is achieved by the key matrix $K$ and value matrix $V$ acquired from the encoder module, and the query matrix $Q$ derived from the previous layer.

To boost the performance of the vanilla self-attention layer, the multihead attention technique, which allows the model to jointly compute the input representation tokens from different subspaces, is widely used.
In particular, given an input vector and the number of heads $h$, all inputs are first linearly mapped into three groups of vectors and packed together: the query group matrix $\{Q_i\}_{i=1}^h$, the key group matrix $\{ K_i \}_{i=1}^h$, and the value group matrix $\{ V_i\}_{i=1}^h$.
Each group contains $h$ vectors with dimensions $d_{q'} = d_{k'} = d_{v'} = d_{model}/h$, and the multiattention process is shown as follows:
\begin{gather}
    \text{MultiHead}(Q', K', V') = \text{Concat}(\text{head}_1, \cdots, \text{head}_h)W^0, \notag \\
    \textbf{where} ~ \text{head}_i = \text{Attention}(Q_i, K_i, V_i),
\end{gather}
where $Q',K', V'$ are the concatenations of the corresponding group matrices $\{Q_i\}_{i=1}^h$, $\{ K_i \}_{i=1}^h$, and $\{ V_i\}_{i=1}^h$, respectively, and $W^0$ is the projection weight.

\subsubsection{Feedforward Network}
A feedforward network (FFN) is a positionwise fully connected network that is applied after the self-attention layer.
It consists of two linear transformation layers with a nonlinear activation function between them:
\begin{equation}
    \text{FFN}(X) = W_2\sigma(W_1X),
\end{equation}
where $W_1$ and $W_2$ are the learned parameters of the two linear transformation layers, and $\sigma$ denotes a nonlinear activation function, such as the Gaussian error linear unit (GELU\cite{GELU}) function or the rectified linear unit (ReLU\cite{ReLU}) function.

\subsubsection{Positional Encoding}
Since the input sequence is ordered and the preceding process is invariant to the position, a positional encoding with dimensionality $d_{model}$ is added to the original input embedding.
The positional embedding of the vanilla transformer is encoded as:
\begin{align}
    PE_{(pos, 2i)} &= \sin (pos / 10000^{2i / d_{\text{model}}}), \\
    PE_{(pos, 2i+1)} &= \cos (pos / 10000^{2i / d_{\text{model}}}),
\end{align}
where $pos$ is the position and $i$ is the dimension.
In this way, each dimension of the positional encoding corresponds to a sinusoid, which makes it easy for the model to learn to apply attention based on relative positions and makes it possible to extrapolate to a longer sequence during the inference process.
Many other positional encoding choices are available, such as learned positional encoding \cite{gehring2017convolutional} and relative positional encoding\cite{shaw2018self}.

\subsubsection{Residual Connections}
To promote the flow of information from the input to the output, a residual connection followed by layer normalization\cite{LN} is applied after each layer in the encoder and decoder modules, shown in Figure \ref{fig:transformer}.
The above process can be represented by the following formula:
\begin{equation}
    \text{LayerNorm}(X + \text{Attention}(X)),
\end{equation}
where $X$ is the input of the self-attention layer.
There are also other variants of residual connections, such as the prelayer normalization (Pre-LN)\cite{baevski2018adaptive,parisotto2020stabilizing}, which places layer normalization before the attention layer, and different normalization algorithms\cite{shen2020rethinking, xu2019understanding}.

%% file: texfile/3.sequencemaking.tex
\section{TRL}
\label{sec:DT}

TRL, whose goal is to leverage the powerful representation ability of the transformer architecture to produce better decision processes in RL tasks, has attracted much attention in recent years.
Considering the success of the transformer architecture in domains where the sequential information process is critical to performance, it is an ideal candidate architecture for partially observable RL problems, where the critical observations often span the entire episode.
Current RL studies often use the LSTM architecture to provide memory to the agent.
These transformer-based methods are able to provide more representative features for agents and environments due to the transformer's superior performance over LSTM, but they only treat the transformer as a tool for architecture enhancement and are still limited by the inherent defects of traditional RL algorithms, such as bootstrapping and the "deadly triad".
Due to the sequential decision process of RL, it is possible to directly involve the transformer architecture in the decision making process and eliminate the limitations of traditional RL frameworks, which is achieved by treating the sequential decision making process of RL problems as a sequence modeling process.
Under the behavior cloning structure, the sequence modeling perspective is able to fully use the long-sequence modeling capability of the transformer and avoid the limitations of traditional RL approaches, such as regularization, conservatism, and the need to discount future rewards.
These methods with simple transformer structures are able to achieve the same performance of traditional RL algorithms that use complex optimization processes, so they have attracted the interest of the RL community.
Since then, many works on decision transformers have been proposed, studying how to further improve performance, how to apply them to other tasks, and the reasons for the success and failure of such methods.
In this section, we introduce the above two TRL methods separately, and name them architecture enhancement (Subsection \ref{subs:AE}) and trajectory optimization (Subsection \ref{subs:TO}) according to the characteristics of the associated algorithms and how their transformer architectures are used, which is visualized in Figure \ref{fig:pipeline}.

\subsection{Architecture Enhancement}
\label{subs:AE}

\begin{figure*}[!t]
    \centering
    \includegraphics[width=7in]{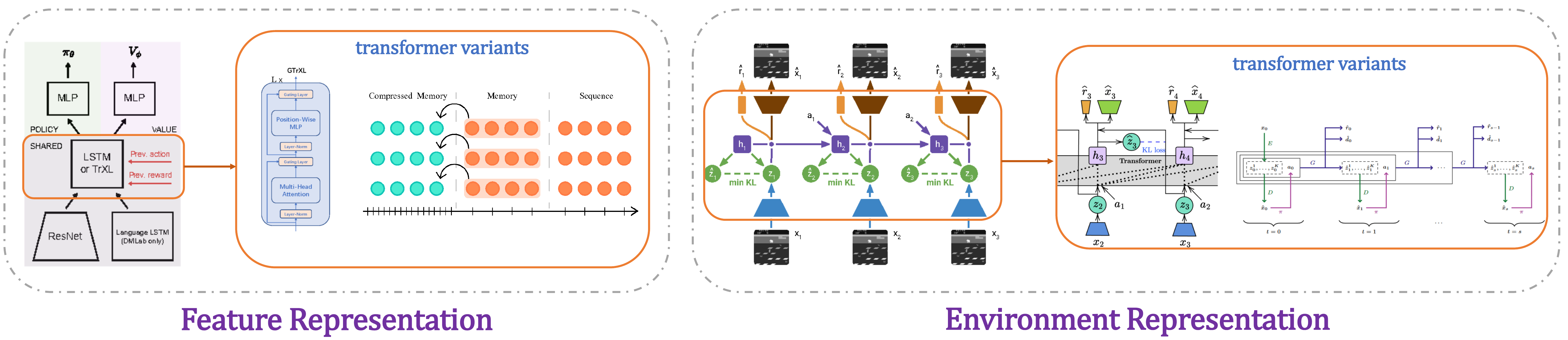}
    \caption{A generic framework for transformer in architecture enhancement. 
    Left: examples of using transformer architecture to enhance feature representations \cite{GTrXL, rae2019compressive}, right: examples of implementing transformer architecture to enhance environment representations \cite{transdreamer, IRIS}.}
    \label{fig:AE}
\end{figure*}

%Traditional RL works often use the transformers as an architectural choice for components.
%
%In this subsection, we first introduce the transformer to provide memory information for the agent networks in domains where the critical observations for any decision often span the entire episode.
%
%Then we introduce the application of transformer in building dynamics and reward functions.
In this subsection, we consider how to apply the powerful transformer structure to RL problems under the traditional RL framework (shown in Figure \ref{fig:AE}).
According to the specific part of the process in which the transformer participates, we further divide this subsection into two parts: feature representation and environment representation.
For feature representation, we mainly introduce the methods where a transformer is used to extract feature representations from a multimodal input, and then the agent utilizes these representations to make decisions through value- or policy-based methods.
For environmental representation, we mainly illustrate how to leverage the transformer architecture for dynamics and reward modeling, which can be used to produce better decision processes with additional planning algorithms.

\subsubsection{Feature Representation}
For a partially observable MDP (POMDP) \cite{POMDP}, conditioning on simple observations is not sufficient for selecting the optimal action; thus, the agent often needs to consider the historical observations.
Instead of directly conditioning the policy on the $N$ most recent observations, using gated recurrent neural networks (RNNs) such as LSTM \cite{LSTM} and gated recurrent units (GRUs) \cite{GRU} to provide memory to the agent is a better choice \cite{impala, kapturowski2018recurrent}.
While recent works have empirically demonstrated that transformers achieve better performance than gated RNNs methods \cite{transformerxl, xlnet} in many domains, a natural question regarding how to incorporate the transformer architecture into the RL domain sparks interest.
Motivated by factored representations \cite{russell2010artificial} and factored MDPs \cite{boutilier2001symbolic}, Loynd et al. \cite{WMG} introduced a working memory graph (WMG), a transformer-based agent that takes a variable number of factor vectors as inputs and outputs their features to an actor-critic network, and the WMG is trained via the A3C algorithm \cite{A3C} with an entropy-regularized policy gradient:
\begin{align}
    \nabla_{\theta} \mathcal{J}(\theta) = \mathbb{E}_{\pi} [\sum_{t=0}^{\infty} \nabla_{\theta} \log \pi(a_t & | h_t ;\theta) A^{\pi}(Obs_t, a_t) \notag \\
    &+ \beta \nabla_{\theta} H(\pi(h_t; \theta))],
\end{align}
where $\pi(a_t | h_t ;\theta)$ is the policy head, $h_t$ is the output of the WMG, $H$ is the entropy and $\beta$ is the hyperparameter.
The WMG also stores the historical information, called Memos, which is the basis of the WMG's shortcut recurrence structure.

\begin{figure}[!t]
    \centering
    \includegraphics[width=2.5in]{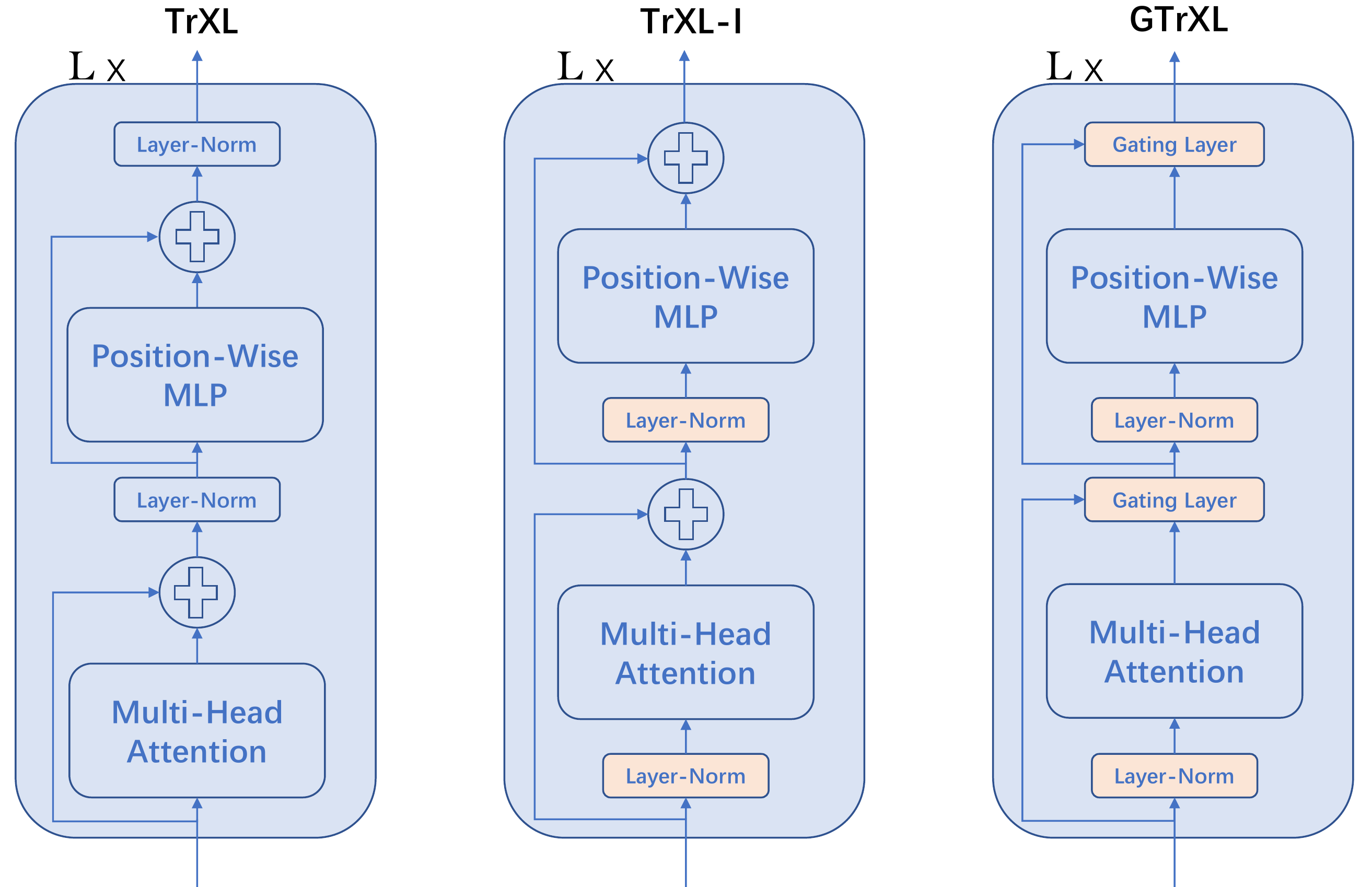}
    \caption{Transformer variants. \textbf{Left:} Standard Transformer(-XL) architecture. \textbf{Center:} TrXL-I puts the layer normalization to the input stream of the submodules. \textbf{Right:} GTrXL replaces the residual connection with the gating layer.}
    \label{fig:GTrXL}
\end{figure}

Concurrently, Parisotto et al. \cite{GTrXL}  tried to leverage the transformer architecture to solve partially observable RL problems.
Different from the WMG, Parisotto et al. modified the standard transformer architecture and produced an XL variant \cite{transformerxl}, called the gated transformer-XL (GTrXL), which substantially improves the stability and learning speed of the transformer(shown in Figure \ref{fig:GTrXL}).
Previous works have shown that the original transformer model suffers from critical instability\cite{mishra2017simple}, and Parisotto et al. also found that it is difficult to optimize the canonical transformer, which results in poor performance.
%
%Though Vinyals et al. \cite{alphastar} succeeded in using transformer in the multi-agent Starcraft 2 environment, the transformer here was applied as the attention over units rather than time.
%
Thus, they replaced the residual connections with gating mechanisms and moved the layer normalization operation onto the ”skip” stream of the residual connections, which is called identity map reordering \cite{nguyen2019transformers, xiong2020layer}.
The gating layer has many options, such as a highway connection \cite{highway}, a sigmoid-tanh (SigTanh) gate \cite{van2016conditional}, and a GRU \cite{GRU}.
In an experiment conducted on DMLab-30 \cite{beattie2016deepmind}, the GtrXL transformer network was used to replace the LSTM-based agents in V-MPO \cite{vmpo} algorithms, and the GRU-type gating layer performed best across all metrics.

Following the GTrXL architecture which functions as a drop-in replacement for the LSTM networks used in RL, many studies have designed more powerful memory transformers for RL.
For example, Rae et al. \cite{rae2019compressive} built on Transformer-XL \cite{transformerxl} and compressed old memories rather than discarding them; then, they used a compressive transformer as a drop-in replacement for the LSTM in the IMPALA algorithms \cite{impala}, as this transformer successfully compresses and makes use of past observations.
Irie et al. \cite{rfwps} noted that linear transformers are equivalent to outer product-based FWPs \cite{schlag2021linear} and that the original FWP formulation is more general than some linear transformers; thus, they augmented the FWP \cite{fwp} with recurrent connections, producing a recurrent FWP (RFWP), and used it as a drop-in replacement for the IMPALA algorithm \cite{impala}, which achieved powerful performance under the standard Atari 2600 setting \cite{bellemare2016increasing} with a small model size and promising scaling properties for a larger model.
Mao et al. \cite{TIT} further designed pure transformer-based networks (TIT) for RL, which is agnostic to the training RL algorithm and provides off-the-shelf backbones for most RL settings.

Compared to LSTM, one disadvantage of a transformer is its significant computational cost, which is important in actor latency-constrained settings.
Thus, Parisotto et al. \cite{ALD} developed an "actor-learner distillation" (ALD) procedure and used a limited-capacity actor to collect data for the learning process of a large-capacity transformer, and then optimized the actor through a policy distillation loss:
\begin{equation}
    L_{ALD}^{\pi} = \mathbb{E}_{s \sim \pi_A} [\mathcal{D}_{KL} (\pi_A(\cdot | s) || \pi_{L} (\cdot | s))].
\end{equation}
For training, they used V-MPO \cite{vmpo} as the main RL algorithm and the IMPALA \cite{impala} distributed RL framework to parallelize the acting and learning processes.

Although it is less able to capture long-range dependencies than a transformer, LSTM has shown a strong ability to effectively capture recent dependencies.
Banino et al. \cite{coberl} proposed a novel agent named contrastive BERT for RL (CoBERL), which combines LSTMs with a transformer as its architecture.
By putting a transformer before the LSTM, the LSTM benefits from the transformer's ability to process long contextual dependencies, and the transformer can reduce the required memory size due to the existence of the LSTM \cite{schwarzschild2021uncanny}.
Banino et al. also combined the paradigm of masked prediction from BERT\cite{bert} with the contrastive approach of RELIC \cite{relic} to improve the performance of agents.
In experiments conducted on the DeepMind Control Suite \cite{tassa2018deepmind} and DMLab-30 \cite{beattie2016deepmind}, they evaluated CoBERL in both on-policy and off-policy settings, where the corresponding RL algorithms were V-MPO \cite{vmpo} and R2D2 \cite{r2d2}, respectively, and optimized a weighted sum of the RL objective and the contrastive loss.

\subsubsection{Environmental Representation}
To achieve high data efficiency, model-based methods have attracted attention.
Recently, several ways to utilize the world model in RL have been proposed, including pure representation learning \cite{schwarzer2020data}, lookahead search \cite{ye2021mastering}, and learning in imagination \cite{hafner2019dream, dreamv2}.
Inspired by the success of the Dreamer agent \cite{dreamv2}, which belongs to the learning in imagination category, Chen et al. \cite{transdreamer} tried to incorporate a more powerful transformer into the RNN-based Dreamer method and proposed a transformer- and model-based RL agent called TransDreamer.
They introduced the first transformer-based stochastic world model, the transformer state-space model (TSSM), by replacing the RNN in the RSSM  \cite{hafner2019dream} with a transformer that is able to support effective stochastic action-conditioned transitions and preserve the parallel computation ability of the transformer.
Empirically, TransDreamer is more powerful than Dreamer on tasks that require long-term and complex memory interactions, which illustrates that the TSSM is better than the RSSM at predicting the future.

In succession, Micheli et al. \cite{IRIS} also studied the ability of a transformer to learn imagination patterns.
For visually complex and partially observable environments, Micheli et al. introduced imagination with autoregression over an inner speech (IRIS), an agent that is trained in an imaginary world, which is generated by a discrete autoencoder and a GPT-like autoregressive transformer.
The autoencoder maps the raw pixels to a much smaller number of image tokens and the transformer simulates the environment dynamics by casting dynamic learning as a sequence generation problem.
During training, the real interaction with the world is used to learn the environment dynamics, the agents only learn from the imaginary world, and the policy learning process opts for learning the objectives and hyperparameters of DreamerV2 \cite{dreamv2}.
In this way, IRIS outperformed a line of sample-efficient RL methods \cite{rainbow, yarats2020image} on the Atari 100k benchmark \cite{atari100k}. 
%
% Similarly,  Ozair et al.\cite{vqm} used the autoencoders to encodes the actions and states of the environment into discrete latent variables and trained a transformer based transition model using the latent variable, which allows the use of Monte Carlo tree search (MCTS \cite{MCTS}) for planning over future actions and observations.
There are also several methods that use autoencoders to encode the actions and states of the environment into discrete latent variables and train a transformer-based transition models using the latent variables.  
Ozair et al.\cite{vqm} used a Monte Carlo tree search (MCTS \cite{MCTS}) to plan future actions and observations, while Sun et al. \cite{plate} adopted a beam search \cite{beamsearch} to reduce performance degradation.

A task-specific reward function is important for guiding agents toward specialists and is meticulously crafted by humans; however, sometimes it is hard to define such a function in the real world or a complex simulation environment.
Fan et al. \cite{minedojo} introduced MINECLIP, which is a contrastive video-language model, to provide reward signals without any domain adaptation techniques for MINEDOJO tasks \cite{minedojo}.
Formally, the MINECLIP reward function can be defined as $\Phi_R: (G,V) \rightarrow \mathbb{R}$, which maps the language goal $G$ and current observation $V$ to a scalar reward.
During training, the function can be optimized via the InfoNCE objective \cite{he2020momentum}, which encourages the behavior in the current snippet to follow the language description.
MINECLIP achieves strong performance and a high-reward snippet conforms to language descriptions, which means that MINECLIP is a good automatic metric for the MINEDOJO simulator.

%% file: texfile/4.sequencemodeling.tex
\subsection{Trajectory Optimization}
\label{subs:TO}

\begin{figure*}[!t]
    \centering
    \includegraphics[width=7in]{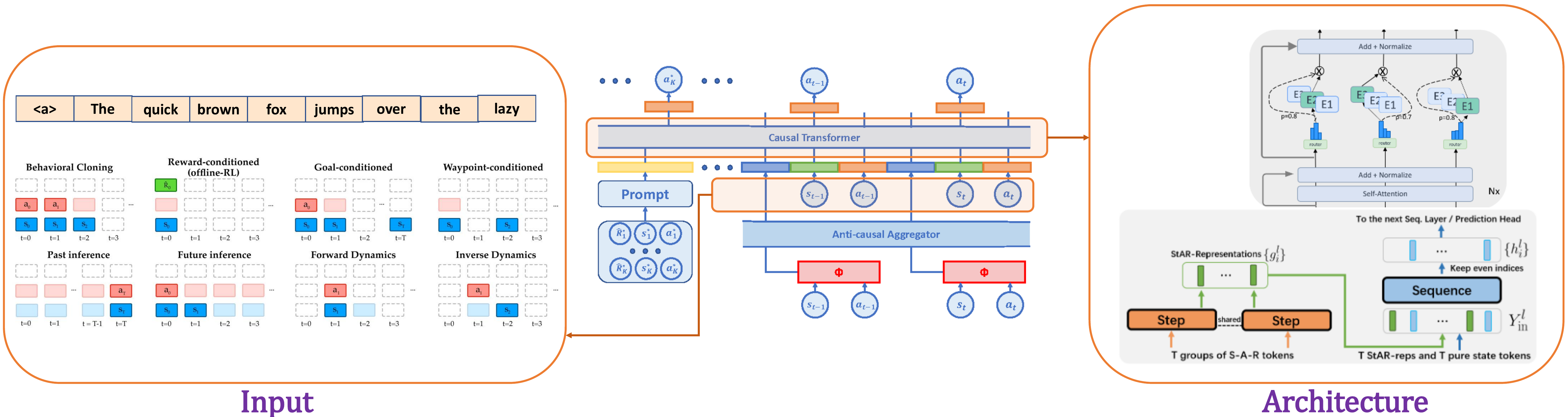}
    \caption{A generic framework for transformer in trajectory optimization. 
    The middle structure is a combination of Prompt-DT\cite{Prompt-DT} and GDT\cite{GDT}. 
    A series of works achieve different purposes by modifying the input form (the left part, mainly for pretraining \cite{chibiT} and generalization \cite{FlexiBiT}) and the structure of the transformer (the right part, mainly to improve performance combining with canonical RL \cite{starformer} and increase generalization \cite{switchtt}) .}
    \label{fig:TO}
\end{figure*}

The RL problem was first considered as a conditional sequence problem by the DT \cite{DT} and TT \cite{TT} algorithms simultaneously. 
Following these pioneering works, a series of new studies have been proposed (shown in Figure \ref{fig:TO}).
In this subsection, we first provide brief introductions to the DT and TT algorithms and introduce some works concerning the inner limitations and optimality conditions of the DT algorithm.
Next, we present some works that combine sequenced modeling methods with traditional RL concepts, which is a natural way to achieve improved performance.
Then we introduce the pretraining mechanism, which is epidemic in NLP and CV, to RL and demonstrate the benefits of incorporating it into the training process.
Finally, we illustrate some algorithms through which a single agent is able to perform well in multiple environments, which may include unseen environments.
%motivated by the success of large scale models in NLP and CV, there are also several works to investigate whether the same strategy adapts to the RL with the help of conditional sequence modeling models.
%
We note that although these methods are sometimes trapped in suboptima, the usage of the transformer architecture and the conditional sequence modeling mechanism makes the RL agents have strong expansibility and generalization.

\subsubsection{Conditioned BC}

\begin{figure*}[!t]
\centering
\subfloat[]{\includegraphics[width=3.5in]{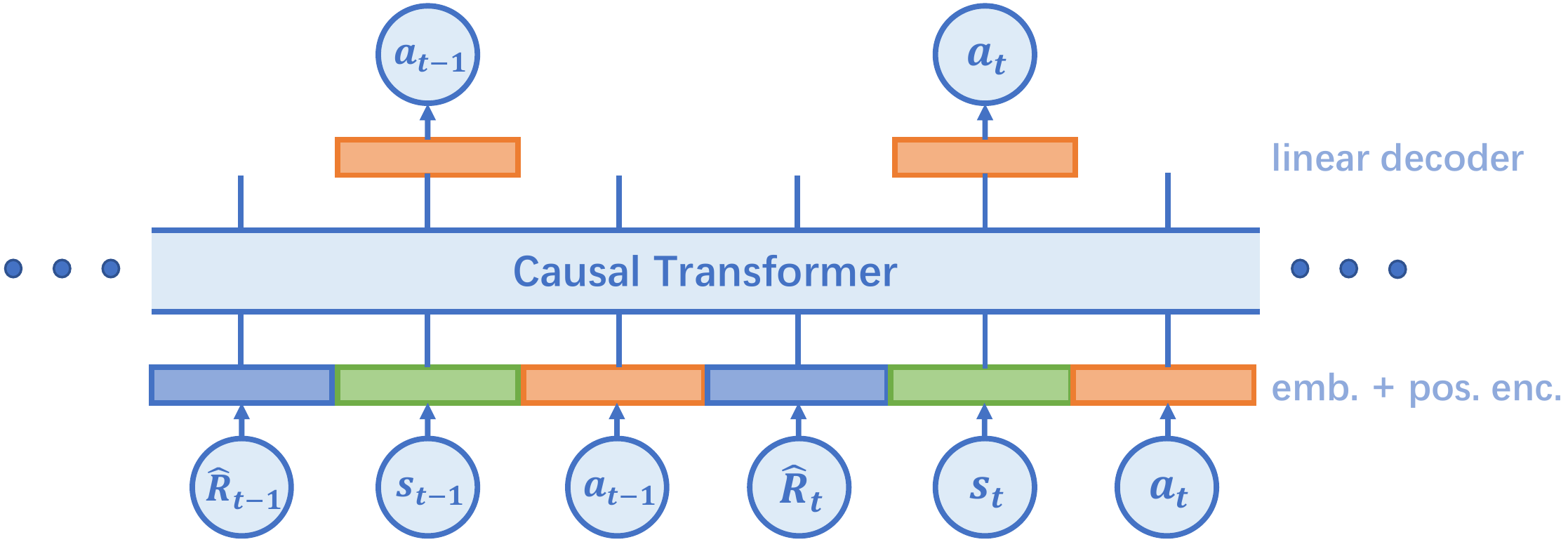}%
\label{fig:DT}}
\hfil
\subfloat[]{\includegraphics[width=3.5in]{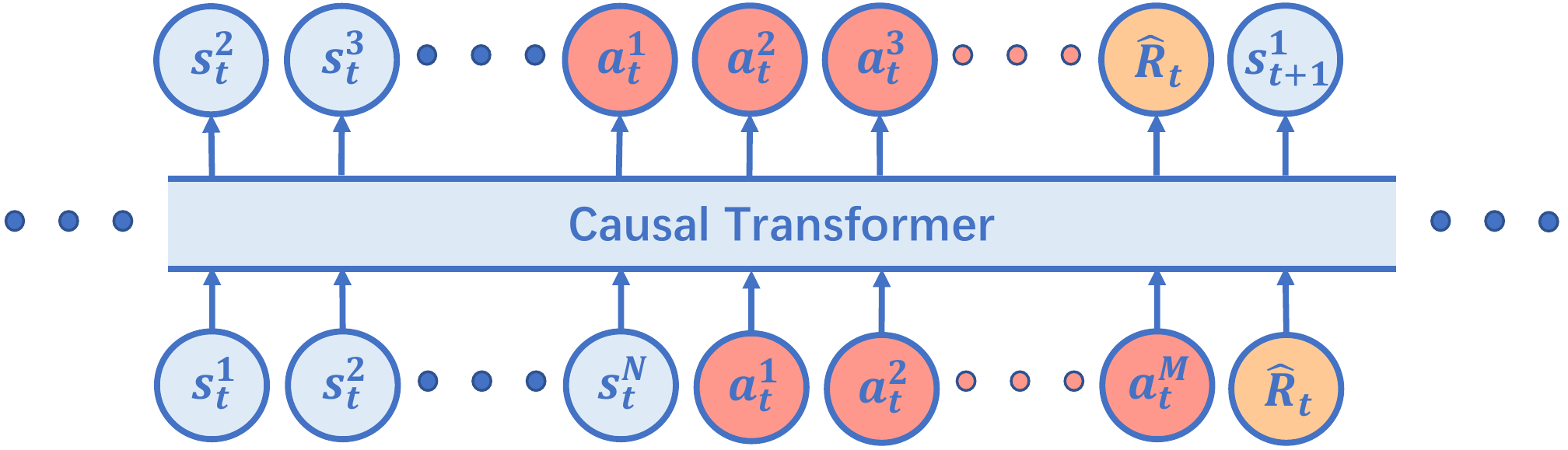}%
\label{fig:TT}}
\caption{(a) Decision Transformer architecture: States, actions, and RTGs are input into the causal transformer which outputs the next timestep actions. (b)Trajectory Transformer architecture: Each dimension of states, actions, and RTGs is input into the causal transformer which outputs the next dimension of states, actions, and RTGs.}
\label{fig_sim}
\end{figure*}

Chen et al. \cite{DT} recently proposed the DT method for offline RL, which treats the policy learning process as a sequence modeling problem (shown in Figure \ref{fig:DT}).
% \begin{figure*}[!t]
%     \centering
%     \includegraphics[width=5in]{figs/DT.pdf}
%     \caption{Structure of Decision Transformer.}
%     \label{fig:DT}
% \end{figure*}
In their formulation, they represented the trajectory as a sequence of state, action, and returns-to-go (RTG) tuples collected at different time steps, which is amenable to autoregressive training and generation:
\begin{equation}
    \tau = (\widehat{R}_1, s_1, a_1, \widehat{R}_2, s_2, a_2, \dots, \widehat{R}_T, s_T, a_T).
\end{equation}
The RTG is the cumulative reward calculated from the current time step until the end of the episode $\widehat{R}_t = \sum_{t'=t}^T r_{t'}$.
Instead of the one-step reward $r_t$, the RTG guides the model to generate actions favoring the future desired returns.
During training, the authors sampled trajectories from offline datasets and minimized the cross-entropy loss for discrete actions or the mean squared error for continuous actions, arguing that predicting the states and RTGs does not help achieve improved performance.
During evaluation, this approach initiates the model with a specific target return based on the desired performance on a given task as well as the environmental state.
Then, the DT selects an action, observes the new state and reward, updates the RTG value by subtracting the observed reward, and repeats this procedure until episode termination.
Although the DT training procedure is similar to behavior cloning, it is more effective than simply performing imitation learning, performs well in sparse settings, and avoids the stability issues related to bootstrapping in long-term credit assignment cases \cite{UDRL, kumar2019reward}.

Concurrently, Janner et al. \cite{TT} proposed the TT, which also uses the transformer architecture to model trajectory distributions, but the TT additionally models the state and reward transitions over the trajectory and discretizes each dimension independently, which incorporates model-based components (shown in Figure \ref{fig:TT}).
In their formulation, assuming $N$-dimensional states and $M$-dimensional actions, the TT represents a trajectory as a sequence of states, actions, and rewards with discretization:
\begin{equation}
    \tau = (s_t^1,s_t^2, \dots, s_t^N, a_t^1, a_t^2, \dots, a_t^M, r_t)_{t=1}^T.
\end{equation}
During training, the TT maximizes the log-likelihood of each token acquired from the sequence in an autoregressive manner.
Once the trajectory distribution is learned, the beam search algorithm\cite{beamsearch} is employed together with the RTG signal to find the reward-maximizing behavior.
Compared to the DT method, the TT predicts all the state, action, and reward tokens and uses an additional beam search algorithm for planning; thus, it can be seen as a model-based method.
In contrast, the DT directly uses the predicted action for planning, making it a model-free method, and supports the possibility that there is no need to include components of traditional RL algorithms for a high-capacity sequence model applied to RL problems.
%\ls{a discussion between DT and TT should be given. what is the difference.}

Although methods such as the DT, which reduce RL to a prediction task and solve it via supervised learning (RvS), have become popular due to their simplicity and strong overall performance on a set of tasks, Paster et al. \cite{esper} noted that in a stochastic environment, the returned reward corresponding to the given trajectory is unstable; therefore, such an approach may fail when simply conditioning a probabilistic model on the desired return and taking the predicted action.
They showed that only when the training goals are independent of environmental stochasticity do these RvS methods consistently achieve optimal policies.
Thus, Paster et al. proposed a method conditioned on the expected returns called ESPER, which is short for environment- and stochasticity-independent representations, to satisfy the above condition.
It first learns a discrete representation $I(\tau)$ acting as a cluster assignment for each trajectory; then, a model learns to predict the average return from the learned representation, which is the condition for an RvS agent to predict the final actions.
Through this method, ESPER is fully conditioned on outcomes that are determined by the actions of the agents and independent of the stochasticity of the environment.

In succession, Brandfonbrener et al. \cite{rcsl} further conducted a rigorous study on the capabilities and limitations of return-conditioned supervised learning (RCSL) algorithms.
They found that the necessary assumptions for an RCSL algorithm to obtain the near-optimal policy are stricter than those required for classic DP methods (e.g., Q-learning).
In particular, RCSL requires nearly deterministic dynamics, knowledge of the target return to be conditioned, and a conditioning value that is consistent with the distribution of the returns in the utilized dataset. 
However, classic DP methods are capable of learning good policies even when the dataset does not contain high-return trajectories and the environment is stochastic.
Paster et al. thought that although RCSL methods such as the DT are simple and have strong performance on many tasks, DP methods still act as general learning paradigms. 
However, the introduction of the transformer structure and the training paradigm of RCSL enables RL agents to have strong expansibility and generalization, so they can be used as general agents with good performance in multitask settings, which is hard for DP methods to accomplish.

\subsubsection{Canonical RL}

Since an offline dataset is always limited and finite, RL methods usually involve an online component, where policies are fine-tuned via task-specific interactions with the environment, and online fine-tuning needs to acquire data via exploration.
Moreover, it has been observed that for standard online algorithms, it is not helpful to access offline data for boosting online performance \cite{nair2020accelerating}; hence, careful consideration is needed to design an overall pipeline.
Zheng et al. \cite{odt} introduced the online DT (ODT), a learning framework for RL that incorporates online fine-tuning into the DT pretraining algorithm.
To balance the exploration-exploitation tradeoff, the ODT explicitly imposes a lower bound on the policy entropy to encourage exploration, which is motivated by the max-env RL framework \cite{levine2018reinforcement}:
\begin{align}
    \min_{\theta} J(\theta) &= \mathbb{E}_{(a,s,g) \sim \tau} [- \log \pi_{\theta}(a | s,g)], \nonumber \\
    &\text{subject to }~~ H^{\tau}_{\theta}[a|s,g] \geq \beta,
\end{align}
where $\beta$ is a hyperparameter, $g$ is the RTG, and the policy entropy $H^{\tau}_{\theta}[a|s,g]$ is defined at the sequence level rather than the transition level, making it different from the entropy used by SAC \cite{SAC}.
Empirically, the ODT attains many extra significant gains during the fine-tuning procedure, which demonstrates its effectiveness and makes it competitive with the state-of-the-art approaches on the D4RL benchmark \cite{d4rl}.
Similarly, Li et al. \cite{PAIR} investigated combining offline behavior cloning with an online PPO \cite{PPO} algorithm and focused on the sparse-reward goal-conditioned robotic manipulation problem.
They adopted the phasic approach by alternating online RL and offline SL with task reduction and an intrinsic reward to achieve improved sample efficiency.

Moving away from online fine-tuning, how to incorporate the Markovian-like inductive bias into a transformer is also important.
Since the sequences of states, actions, and rewards generally have strong connections due to their potential causal relations, excess information or diluted essential relations may be encountered when simply using a transformer that attends to all tokens.
It is also difficult to learn the Markovian-like dependencies between tokens from scratch due to the weak relationships between nonadjacent tokens \cite{TT}.
To alleviate such issues, Shang et al. \cite{starformer} proposed the state-action-reward transformer (StARformer), which consists of two interwoven components: a step transformer that learns the local Markovian representations by self-attending a state-action-reward triple, and a sequence transformer for modeling long-term dependencies.
Note that the StARformer can directly operate on a stepwise reward without performance degradation, while it is important for the DT to carefully design the RTG in the inference stage.

Afterward, Laskin et al. \cite{AD} indicated that the policies learned by sequence modeling methods do not incrementally improve based on additional interaction with the environment.
They hypothesized that the reason these methods cannot be improved through trial and error is that the training data do not yield learning progress: the data either involve no learning (e.g., they are collected by fixed expert policies) or exhibit learning but with a context size that is too small to capture the policy improvement.
Thus, Laskin et al. presented algorithm distillation (AD), which learns an in-context policy improvement operator on the learning history of an RL algorithm.
An offline training dataset needs to be sufficiently large (i.e., across episodes) to capture the policy improvement; then, a causal transformer uses the learning history as its context to model the actions, and the learning loss function is as follows:
\begin{equation}
    \mathcal{L}(\theta) := - \sum_{n=1}^N \sum_{t=1}^{T-1} \log P_{\theta} (A = a_t^{(n)} | h_{t-1}^{(n)}, o_t^{(n)}). 
\end{equation}
In particular, during training, AD first randomly samples a multiepisodic subsequence of length $c$ and then autoregressively predicts the actions with a transformer.
During an evaluation in the DMLab Watermazer environment \cite{morris1981spatial}, AD stores the sequential transitions in a context queue when unrolling the transformer in the environment.
Through this method, AD can learn new tasks in an entirely in-context manner without updating its network parameters since all updated information is contained in the historical context.

For offline RL tasks, two major limitations that might cause suboptimal training are encountered.
One is that the coverage of an offline dataset is limited, and coverage of all state-action-reward transitions cannot be guaranteed \cite{fujimoto2019off, qin2021neorl}; the other is the limited amount of available training data, which may deteriorate the performance of the constructed model \cite{lan2019albert}.
In addition to training an additional model for learning the environment dynamics to expand the data coverage, Wang et al. \cite{BooT} proposed a novel algorithm called the bootstrapped transformer (BooT), which first generates data via the learned model itself and then uses the generated data to further train the model for bootstrapping.
Due to the capability of the DT architecture, the data generated by the model are consistent with the original offline data, thus prevent overfitting during the training procedure \cite{liu2021augmenting}.
To stabilize the bootstrap training process, BooT only conducts trainings on the generated trajectories with the highest confidence scores to prevent training with inaccurately generated data, where the confidence is defined as:
\begin{equation}
    c(\tau) = \frac{1}{T'(N+M+2)} \sum_{t=T - T' +1}^T \log P_{\theta} (\tau_t | \tau_{<t}),
\end{equation}
where $N~ \text{and}~ M$ are the dimensions of the state and action (similar to the TT), respectively.
This approach has demonstrated great effectiveness in extensive experiments, which were based on the D4RL offline dataset \cite{d4rl}, but it requires a long training time due to the pseudo data generation step.

Similarly, Yamagata et al. \cite{QDT} noted that if an offline dataset only contains suboptimal trajectories, the DT method fails to find the optimal method, which means that it lacks the stitching ability, i.e., the ability to learn the optimal policy from suboptimal trajectories, possessed by conventional RL approaches (e.g. Q-learning).
Thus, Yamagata et al. proposed the Q-learning DT (QDT), which leverages Q-learning to improve the DT by improving the quality of the utilized dataset; the QDT relabels the RTG tokens in the offline training data.
In particular, since the conservative Q learning (CQL) \cite{CQL} framework learns lower bounds of the true Q-function values, it only replaces an RTG value when it is lower than the lower bound calculated by the learned value function and replaces all RTG tokens prior to the changed RTG to maintain consistency: $R_t = r_t + R_{t+1}$.
The QDT is able to perform well across all environments (e.g., Maze2D, Hopper, HalfCheetah, and Walker2d in D4RL \cite{d4rl}), while both Q-learning and the DT have poor performance in some environments, which means that the QDT combines the strength of these approaches.

\subsubsection{Pretraining}
With regard to sequence modeling, large pretraining models achieved impressive successes in NLP \cite{bert} and CV \cite{dosovitskiy2020image} tasks since pretraining is an essential technique for alleviating the high computation costs induced by using expressive models such as transformers.
Treating RL as a sequence modeling problem allows us to take advantage of the pretraining technique; however, there is a lack of large off-the-shelf RL datasets for pretraining \cite{cobbe2020leveraging, yu2020meta}.
Instead, Reid et al. \cite{chibiT} demonstrated that conducting pretraining on natural language can also provide drastic convergence speed and final policy performances improvements in offline RL tasks that are unrelated to language (e.g., MuJoCo \cite{mujoco}), providing a unified view of the sequence modeling domain.
To better adapt a pretrained language model to model trajectories, Reid et al. used a similarity-based objective to maximize the similarity between the language embeddings and the input trajectory representations:
\begin{equation}
    \mathcal{L}_{cos} = - \sum_{i=0}^{3N}\max_j \mathcal{C}(I_i, E_j),
\end{equation}
where $E=[E_1, \dots, E_V]$ are the language embeddings with a vocabulary size of $V$, $I=I_1, \dots, I_{3N}$ are the trajectory input tokens, and $\mathcal{C}(z_1, z_2) = \frac{z_1}{||z_1||_2} \cdot \frac{z_2}{||z_2||_2}$ is the cosine similarity function, respectively.
Reid et al. also showed that visual initialization does not yield a performance improvement and even has a negative effect, which illustrates the large difference between image modeling and trajectory modeling.
%
%Above all, the sequence modeling treatment does away with many complexities with RL and allows to take advantage of techniques popularized in sequence modeling tasks for RL.

Following this work, Li et al. \cite{LID} further noted that using pretrained language models as general scaffolds improves combinatorial generalization in policy learning, especially for out-of-distribution tasks.
They also performed several ablation experiments on BabyAI \cite{babyai} and VirtualHome \cite{virtualhome} and showed that there is no need to convert the historical states and actions to natural language strings to benefit from language pretraining; however, sequential input encoding and language pretraining are important.
Takagi \cite{Takagi2022on} empirically investigated the impact of conducting pretraining on data with different modalities and studied what information is important when performing pretraining on language data.
He showed that the transformer obtained after pretraining largely changes its representations and does not pay attention to the downstream tasks. 
He also noted that the reasons pretraining with image data attains poor performance \cite{chibiT} include the large gradients and gradient clipping.
The reason why pretraining with language data performs well may be that it makes the transformer obtain context-like information, which is then used to solve the downstream tasks.

Concurrently, Liu et al. \cite{MaskDP} randomly masked state and action tokens and reconstructed missing data, and they focused on solving downstream tasks via masked pretraining.  
They proposed masked decision prediction (MaskDP) and found that conducting masked pretraining on unsupervised data leads to better general agents.
They also found that the mask ratios produced during pretraining are important for downstream task performance since the states and actions in one sequence are highly correlated temporally.

Yang et al. \cite{ACL} also studied the effect of pretraining on downstream tasks: what unsupervised objectives are able to improve the performance of downstream tasks, and what components of the objectives are most important?
They leveraged the D4RL \cite{d4rl} offline dataset and mainly considered three categories for downstream tasks: imitation learning with few data, offline RL, and online RL.
They used a transformer-based architecture, which was trained to predict a randomly masking subset with a corresponding contrastive loss for pretraining, and then combined different algorithms for different downstream tasks, such as BRAC \cite{brac} for offline RL and SAC \cite{SAC} for offline RL.
The results obtained on Gym-MuJoCo from D4RL \cite{d4rl} showed that the use of pretraining with an unsupervised learning objective could improve the performance of policy learning, and the best representation learning objective for pretraining depends on the given downstream task, which means that no single objective is optimal for all downstream tasks.

\subsubsection{Generalization}
It is difficult for traditional RL agents to build large-scale general agents by conducting training on massive datasets since they tend to use small models to solve single tasks or different tasks in one dynamic \cite{cuccu2018playing, bastani2018verifiable}.
Initially, for the Atari suite \cite{ALE}, deep Q-learning \cite{mnih2013playing} and actor-critic methods \cite{mnih2016asynchronous} tried to play all games with one algorithm, but they both required separate training processes for each game.
Then, later works tried to learn a single neural network agent to play multiple Atari games simultaneously \cite{impala, parisotto2015actor, rusu2015policy}.
Lee et al. \cite{multigame_DT} also tried to learn a single agent for playing multiple Atari games.
They investigated several approaches, including DT-based sequence modeling approaches \cite{DT}, online RL \cite{DQN}, offline temporal difference methods \cite{CQL}, contrastive representations \cite{oord2018representation} and behavior cloning \cite{pomerleau1991efficient}, and found that DT-based sequence modeling offers the best scalability and performance.
Since the given training dataset usually contains expert and nonexpert trajectories, simply imitating the data cannot guarantee the production of expert actions all the time.
Inspired by the discriminator-guided generation problem in language models \cite{krause2020gedi, ouyang2022training}, Lee et al. proposed an inference time method that assumes a binary classifier $P(\text{expert}^t | ...)$ and used Bayes' rule to generate expert-level returns and actions:
\begin{equation}
    P(R^t | \text{expert}^t, \dots) \propto P_{\theta}(R^t | \dots) P(\text{expert}^t | R^t, \dots),
\end{equation}
where the expert is proportional to the future return $P(\text{expert}^t | R^t, \dots) \equiv \exp (\kappa R^t)$, similar to probabilistic inference \cite{shachter1988probabilistic, toussaint2009robot}.
An experiment on 46 Atari games with an existing offline dataset \cite{agarwal2020optimistic} showed that the scaling rule can also be adapted to RL: the larger the size of the model is, the better its performance and the faster it adapts to new games by fine-tuning.

Concurrently, Reed et al. \cite{Gato} aimed to build a transformer-based general agent based on offline data.
They proposed an agent named Gato to work as a multimodal, multitask, multiembodiment general policy that trains on the widest variety of relevant data possible.
To process multimodal data through one agent, they first serialized all data into a flat sequence of tokens and then used the transformer model to output the distribution over the next discrete token through the following training loss:
\begin{equation}
    \mathcal{L}(\theta) = -\sum_{b=1}^{|\mathcal{B}|} \sum_{l=1}^L m(b,l) \log p_{\theta} (s_l^{(b)} | s_1^{(b)}, \dots, s_{l-1}^{(b)}),
\end{equation}
where $\mathcal{B}$ denotes the training batch sequences, and the masking function $m(b,l)=1$ if the token at index $l$ is from the text or logged action of an agent.
Note that the training dataset exclusively comprises near-optimal agent experience in both simulated and real-world environments, and Gato needs the extra information prompted by expert trajectories to infer the next token during evaluation \cite{sanh2021multitask, wei2021finetuned}, making it different from Lee et al.'s work.

Similarly, Lin et al. \cite{switchtt} used high-capacity sequential models to solve the multitask RL problem and indicated three significant algorithmic challenges: the high computational cost and model capacity requirements, the negative impact on single-task training caused by parameter sharing, and the poor performance caused by the Monte Carlo value estimator, which suffers from poor sample complexity.
They proposed the switch trajectory transformer (SwitchTT), which enhances two features of the TT method.
First, it exploits a sparsely activated model by replacing the FFN layer with a switch layer consisting of multiple FFN layers and a router function to route the input to a specific expert \cite{fedus2021switch}.
This modeling process enables efficient computation and promotes parameter sharing in multitask learning settings.
Second, it models the value distribution of a trajectory rather than the expected value to mitigate the effect of the Monte Carlo value estimator and incorporate the uncertainty into the reward.
In practice, SwitchTT uses the categorical distribution and minimizes the cross-entropy loss between the labeled class and the discrete distribution, and in the planning phase, it uses the weighted sum of the discrete atoms as the value of each trajectory:
\begin{equation}
    V(x) = \sum_{i=1}^{N} p_i(x) z_i,
\end{equation}
where $z_i$ is the discrete value and $p_i$ is the corresponding probability given by the model.
The multitask training dataset is composed of the replay buffer generated by the PPO \cite{PPO} algorithm, which contains both expert and nonexpert trajectories.
The experimental results obtained on 10 tasks in the gym minigrid environment (e.g., FourRoom, DoorKey, and KeyCorridor) showed that SwitchTT achieves comparable performance to that of the TT while saving 90\% of the training time.

\begin{figure*}[!t]
\centering
\subfloat[]{\includegraphics[width=3in]{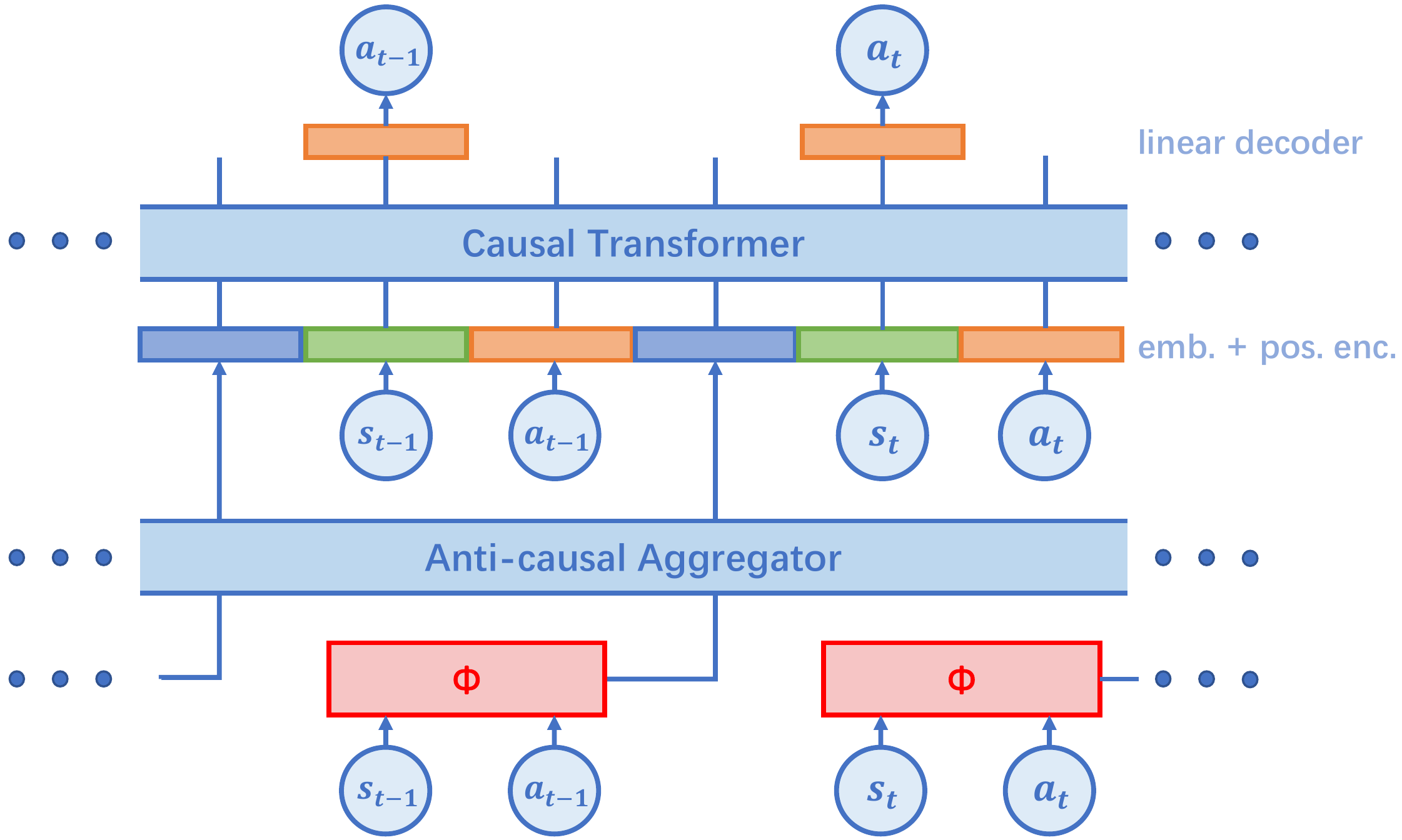}%
\label{fig:GDT}}
\hfil
\subfloat[]{\includegraphics[width=3.5in]{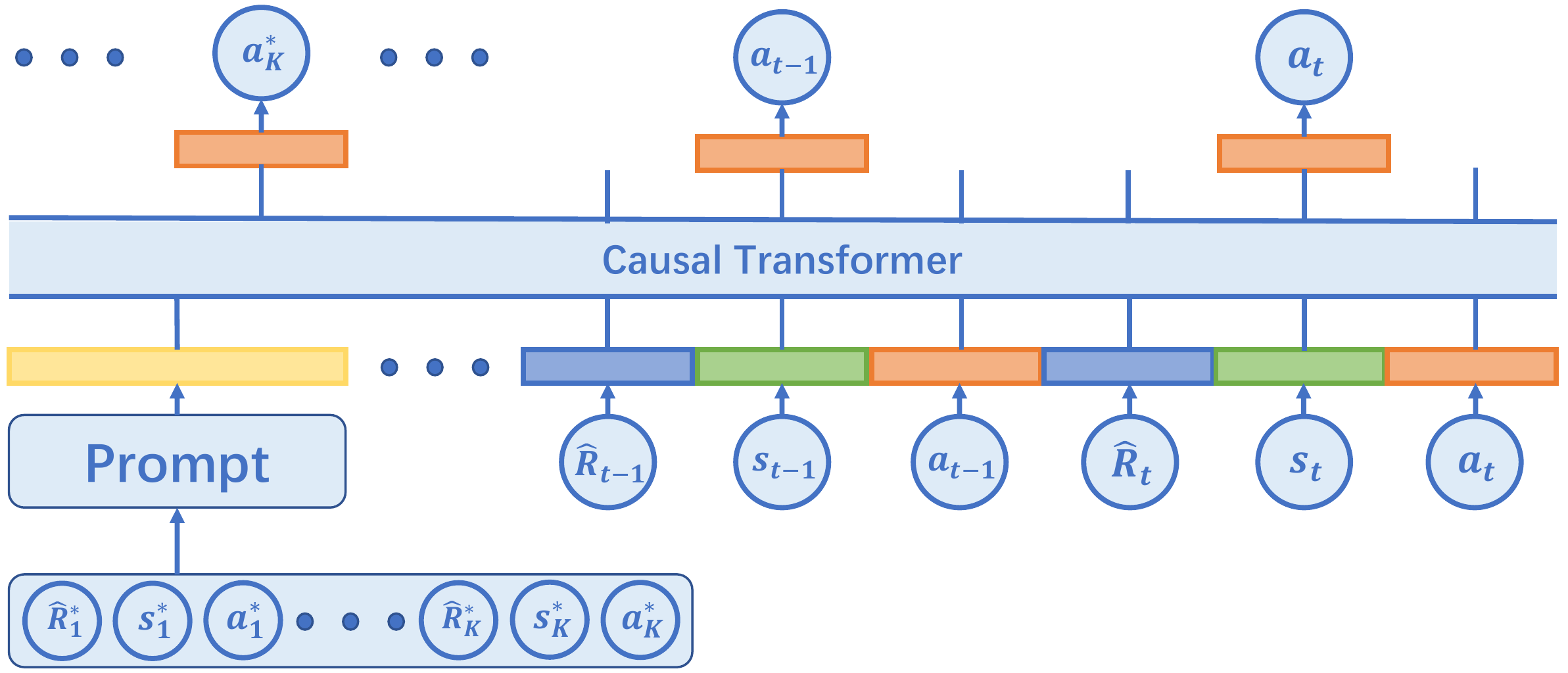}%
\label{fig:Prompt-DT}}
\caption{(a) Structure of Generalized Decision Transformer, which is a generalization of the DT structure including an anti-causal aggregator and a feature function $\Phi$. (b) Structure of Prompt-DT, which takes the trajectory prompt augmentation and the history sequence as input, and auto-regressively predicts the actions corresponding to each state.}
\label{fig_sim}
\end{figure*}

Furthermore, Furuta et al. \cite{GDT} demonstrated that many algorithms (e.g., hindsight multitask RL \cite{li2020generalized, eysenbach2020rewriting} and upside-down RL \cite{srivastava2019training, kumar2019reward}) conditioned on future trajectory information can be unified through hindsight information matching (HIM), which trains policies to output future tokens while satisfying some statistics with respect to future state information.
They defined the information matching (IM) problem as learning a conditional policy $\pi(a | s, z)$ by minimizing the divergence between the statistics of the generated trajectory and the desired information $z$:
\begin{equation}
    \min_{\pi} \mathbb{E}_{z \sim p(z), \tau \sim \rho_{z}^{\pi}(\tau )} [D(I^{\Phi}(\tau), z)],
\end{equation}
where $I^{\Phi}(\tau)$ denotes the information statistics of the trajectory $\tau$. 
In practice, it can be the identity function, the reward function $r(s,a)$, and so on.
Furuta et al. accordingly introduced a generic template, the generalized DT (GDT), which adjusts its feature function and anti-causal aggregator to adapt to different choices of $I^{\Phi}(\tau)$ (shown in Figure \ref{fig:GDT}).
% \begin{figure}[!t]
%     \centering
%     \includegraphics[width=3in]{figs/GDT.pdf}
%     \caption{Structure of Generalized Decision Transformer, which is a generalization of the DT structure.}
%     \label{fig:GDT}
% \end{figure}

In succession, Carroll et al. \cite{FlexiBiT} provided a unified view of different tasks, such as BC, goal- and waypoint-conditioned imitation \cite{ding2019goal, rhinehart2019precog}, offline RL, dynamics prediction \cite{ha2018world, christiano2016transfer}, and initial state inference \cite{shah2019preferences}, as sequence masking tasks.
They proposed a FlexiBiT framework that is trained on a variety of inference tasks by a single agent.
Since each task corresponds to a unique masking scheme, to obtain a general agent that performs well on any arbitrary sequence inference, FlexiBiT is trained by the random masking scheme which masks each token independently with a probability of $p_{\text{mask}} \sim \text{Uniform}(0,1)$.
Such a training method enables FlexiBiT to perform well on any inference task, which is based on the mini-grid environment framework \cite{chevalier2018minimalistic}, as the generalist agent, and even outperforms specialized models that only train on the given tasks.
This means that single-task models built via unified multitask training further achieve improved performance.

Another difficulty of RL generalization involves generalizetion to unseen tasks.
A common method is to perform offline meta-RL (e.g., MAML \cite{MAML} or MACAW \cite{MACAW}) to achieve quick adaptation via algorithmic design.
Inspired by the prompt-based framework for adaptation in NLP \cite{liu2021pre, brown2020language}, Xu et al. \cite{Prompt-DT} proposed the prompt-based DT (Prompt-DT) which focuses on the power of 
inductive architecture bias.
Instead of utilizing a text description as a prompt, which requires much human labor to annotate \cite{shridhar2020alfred}, Xu et al. adopted trajectory segments as prompts and forced the agent to imitate these demonstrations without fine-tuning (shown in Figure \ref{fig:Prompt-DT}).
Further experiments (e.g., Cheetah-dir \cite{MAML}, Ant-dir \cite{mitchell2021offline}, and Dial \cite{yu2020meta}) showed that given few-shot trajectories, the Prompt-DT is able to beat strong offline meta-RL baselines and can be generalized to out-of-distribution tasks.

Similarly, Boustati et al. \cite{transfer-DT} focused on training agents to make them robust to environment dynamics changes via causal reasoning.
They enriched the training dataset with counterfactual trajectories to boost the agent's adaptability to structural changes.
Mathematically, followed by causality \cite{pearl2009causality}, Boustati et al. modeled the transitions as $p(s' | s, a, \Phi)$ with $\Phi$ denoting the structural features of the environment, and one can generate a set of counterfactual trajectories by intervening in $\Phi$.
To perform training on more effective counterfactual trajectories, they used the ATE measure to order these trajectories:
\begin{equation}
    \text{ATE}[\phi_{CF}] = \sum_{t=1}^T \mathbb{E}_{p_{\text{CF}}(\tau)} [r_t] - \mathbb{E}_{p_{\text{Source}}(\tau)}[r_t].
\end{equation}
%
%However this generation method needs to know the structural equation model of the environment which is hard for most problems.
The empirical results obtained on the gym\_minigrid environment suite \cite{chevalier2018minimalistic} proved the efficacy of this method.

Some works have focused on generalization from other perspectives, such as the AnyMorph \cite{anymorph} approach that trains a policy to generalize to new agent morphologies without fine-tuning, the AttentionNeuron \cite{AttentionNeuron} method that deals with sudden random input reordering while performing a task, etc.

%% file: texfile/5.application.tex
\section{Applications of TRL}
\label{sec:application}

TRL has many specific applications. 
Here, we mainly introduce four popular environments, including robotic manipulation, TBGs, vision-language navigation, and autonomous driving.
%
%These fields usually need to process multi-modal input in a complex environment, so the transformer structure are fully utilized. 
In each of these fields, agents usually need to process multimodal inputs in a complex environment, making them suitable situations for fully utilizing the capabilities of the transformer architecture.
%
%Next, we will describe the modeling methods \ls{modeling methods? Chinese-English} and corresponding solutions in each field in turn.
Below, we describe the problem formulation of each field and present corresponding solutions with the help of the transformer architecture.

\subsection{Robotic Manipulation}
For one-shot robotic imitation learning, it is important to consider which pattern to use to represent the goal/intention derived from visual demonstrations and how to incorporate it into the policy network with the current observations.
Dasari et al. \cite{T-OSIL} explored task-driven features to produce a one-shot learning strategy and used a transformer (a nonlocal self-attention block \cite{non-local} in the multihead version) to extract relational features as the input state vector for the policy function.
The transformer takes the features derived from both the demonstrations and current observations via 
the ResNet \cite{ResNet} process as inputs, which is beneficial for preserving the spatial and temporal information in the attention module.
However, at the test time, the transformer features fail to weigh the important details, and Dasari et al. further added an unsupervised inverse dynamics loss that forces the transformer to model the underlying dynamics.
In particular, the context and observations are first randomized and then passed to the representation model to generate features $\Tilde{\phi}_t$, which are used to predict a discretized logistic mixture distribution \cite{salimans2017pixelcnn++}.
The inverse loss is:
\begin{align}
    \mathcal{L}_{inv} (\mathcal{D}, \theta) &= - \ln (\sum_{i=1}^k \alpha_k (\Tilde{\phi}_t, \Tilde{\phi}_{t+1}) \notag \\
    & \cdot \text{logistic}(\mu_i(\Tilde{\phi}_t, \Tilde{\phi}_{t+1}) , \sigma_i (\Tilde{\phi}_t, \Tilde{\phi}_{t+1}) ).
\end{align}
The results obtained in multiagent MuJoCo environments \cite{mujoco} showed that both the network design and the inverse loss can help policies perform better at test time, and further ablations showed that transformer models are also more powerful than LSTM with a traditional "embedding-level" attention mechanism.
In succession, Zhao et al. \cite{MOSAIC} noted that prior works \cite{T-OSIL, yu2018one} tended to assume very strong similarity between training and testing in one-shot imitation learning (OSIL) settings, and they introduced a new simulated robotic manipulation benchmark and a new method called MOSAIC.
Compared to T-OSIL \cite{T-OSIL}, MOSAIC uses a self-attention-based policy network but adopts the temporal contrastive loss objective \cite{oord2018representation} which is motivated by the fact that the representations of two adjacent frames should be as similar as possible while those of nonadjacent frames should be far away, i.e.,
\begin{equation}
    \mathcal{L}_{Rep} = \log \frac{\exp(q^TWk_+)}{\exp(q^TWk_+) + \sum_{i=1}^{F-1}(q^T W k_i)},
\end{equation}
where $q ~ \text{and}~ k_+$ are the anchor and its positive, respectively, $W$ is the projection matrix and $F$ is the total frame count.
Experiments conducted on Robosuite \cite{zhu2020robosuite} and MetaWorld \cite{yu2020meta} showed that MOSAIC has powerful learning efficiency and performance and is able to learn a promising general multitask policy via fine-tuning.

\begin{figure}[!t]
    \centering
    \includegraphics[width=3in]{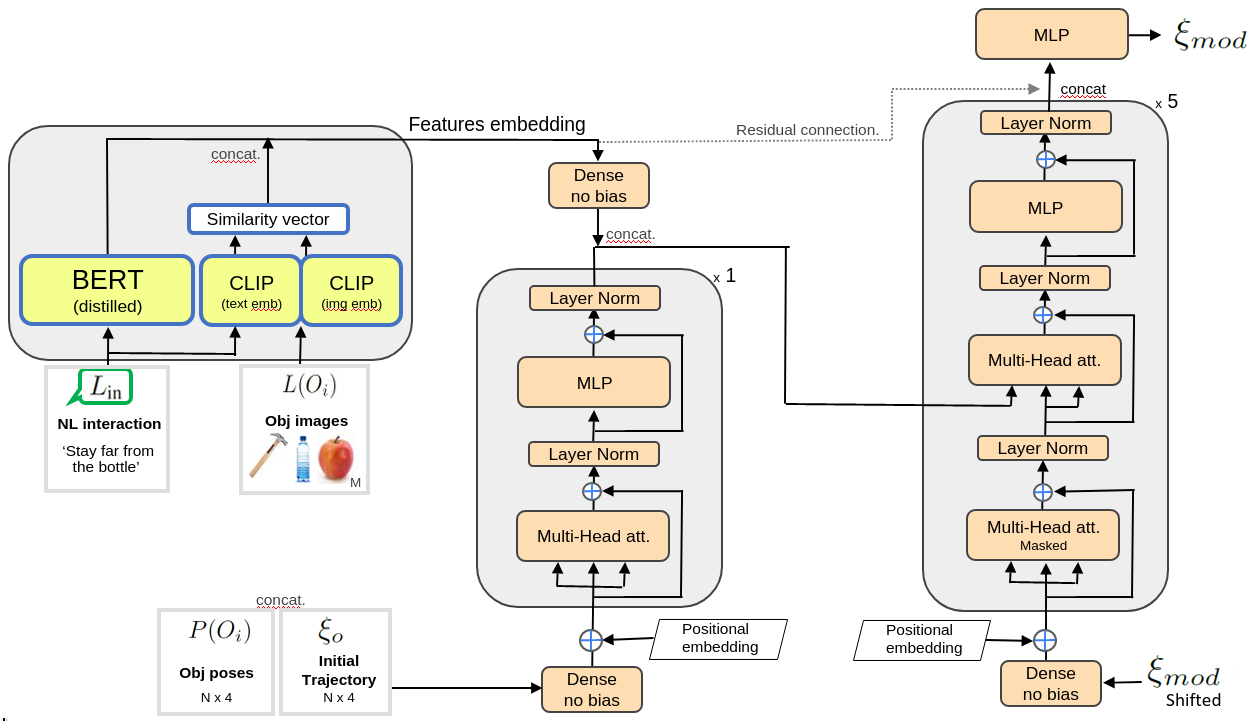}
    \caption{The framework of LATTE (image from \cite{LaTTe}), which includes a language and image encoder, a geometry encoder, and a multi-modal transformer decoder.}
    \label{fig:LATTE}
\end{figure}

In the human-robot interaction setting, natural language is one of the most intuitive ways to express human intent to a robot.
However, it is not an easy task to guide the robot to generate the target trajectory under human instructions and commands.
Rather than defining rigid templates with a static set of actions and commands, Bucker et al. \cite{bucker2022reshaping} provided a flexible language-based interface where a user can reshape the existing robotic trajectory according to the human commands.
By treating the trajectory generation process as a sequence modeling problem, it is natural to leverage the powerful transformer language model.
Specifically, Bucker et al. first employed large pretrained language models to encode human instructions, providing rich semantic representations, and a pretrained language model can also lower the requirement regarding the number of training examples.
Then, they jointly aligned the geometrical trajectory data with language embeddings through the use of a multimodal attention mechanism.
Experiments showed that the combination of pretrained large language models with multimodal transformers could create more intuitive interfaces between robots and humans than traditional methods such as kinesthetic teaching and programming-based obstacle avoidance.
Afterward, Bucker et al. \cite{LaTTe} further enhanced the architecture to include an actual image encoder \cite{CLIP} and modeled the trajectory in 3D space with velocity rather than just XY movement (shown in Figure \ref{fig:LATTE}).
%They evaluated it in a more diverse scenarios setting, such as the aerial and legged robots.
%
Similarly, Sharma et al. \cite{sharma2022correcting} also used a model with the transformer architecture for encoding instructions and to create a semantic cost map that guides the planner to produce a trajectory.

Furthermore, Guhur et al. \cite{Hiveformer} focused on designing generic instruction-following agents that produce actions other than modifying existing trajectories and noted that recent works mainly relied on current observations \cite{shridhar2022cliport, jang2022bc}, which may cause a problem in sequential tasks requiring object tracking and remembering previous actions; thus, they introduced a new model, the Hiveformer, which jointly models human instructions, multiple views and historical information via a multimodal transformer.
The Hiveformer simultaneously encodes the input instructions, combining past and current visual observations and proprioception, and predicts 7-degree-of-freedom (DoF) actions.
Experiments conducted on RLBench \cite{rlbench} demonstrated the strong generalization ability of the Hiveformer, whose historical input contributes to long-term tasks and whose multiview input contributes to tasks requiring high precision or those involving occluded objects.
Similarly, Shridhar et al. \cite{PERACT} used a Perceiver transformer \cite{jaegle2021perceiver} to exploit 3D structure voxel patches and language goals to predict 6-DoF actions.

For task-specific settings, the robot needs to know the task constraints and the dynamic environments.
Jain et al. \cite{TTP} thought that a robot should also adapt to users' different preferences and proposed a prompt-situation architecture, called TTP, which is able to generalize to unseen preferences at test time with a single user demonstration.
Assuming that the demonstrations implicitly contain the task structure and preference, TTP takes one demonstration trajectory and the current state as input and learns to imitate the preference in the demonstrations:
\begin{equation}
    \min_{m \sim M, \tau \sim D_m, (S,a) \sim D_m} \mathcal{L}_{CE} (a, \pi(S, \psi(\tau)))
\end{equation}
where $m$ is the preference, $D$ is a multipreference dataset and $\tau$ is the prompt containing state-action pairs.
TTP has successfully validated its properties in both a simulated kitchen environment and the real-world Franka Panda robotic arm.
%
%While Sun et al. \cite{plate} try to plan a goal-conditioned procedure from the videos where the robot should directly learn the structure and plannable state and actions from the video, and proposed the 

A common use of transformers in the robotic manipulation domain is to serve as feature extraction modules for one or more modalities simultaneously.
Many works are still designing transformer architectures to better extract representations from visual/language/text inputs for downstream tasks, such as grippers \cite{han2021learning, jangir2022look}, legged locomotion \cite{yang2021learning}, and dual-arm robots \cite{kim2021transformer}, and these methods are mainly trained on BC patterns.

\subsection{TBGs}
TBGs are interactive simulations where an agent plays a game by processing text observations and acting out text commands.
Formally, TBGs can be seen as POMDPs; the agents do not know the end goal from the start and earn sparse rewards by completing a sequence of subgoals.
The difficulty of a TBG mainly depends on the environment's complexity, the game length, and the verbosity \cite{textworld}.
Several frameworks also promote the development of TBGs, such as TextWorld \cite{textworld}, LIGHT \cite{LIGHT}, Jericho \cite{Jericho} and TextWorld with QA \cite{yuan2019interactive}.

% It is important for agents to overcome bottleneck states which is dependent on some items or locations.
% %
% The quests in text games can be modeled in the form of a dependency graphs: $\mathcal{D}=\langle V, E \rangle$, which are directed acyclic graphs (DAGs), and each vertex is tuple $v = \langle s_l, s_i, r(s) \rangle$ where $s_l, s_i, r(s)$ are the location dependencies, inventory dependencies and reward respectively, each directed edge means the originating state satisfies the requirements $s_i$ and $s_l$ of the terminating state. 
% %
% When $\mathcal{D}$ is topologically sorted into levels $L = \{ l_1, \dots, l_m$, the bottleneck states can be define as:
% \begin{align}
%     \mathcal{B} = \{b: & (|l_i|=1, b \in l_i, V) \notag \\
%     & \wedge (\exists s \in l_j ~s.t. ~ (j > i \wedge r(s) \neq 0))   \}.
% \end{align}
% However most methods \cite{yin2019comprehensible, kga2c} fail to consistently clear these bottleneck states since they have long-range dependencies and are usually not rewarded.
%
Standard RL algorithms are poor at effectively exploring entire spaces since most games have over a billion possible actions per step \cite{Jericho, kga2c}.
Ammanabrolu et al. \cite{ammanabrolu2018playing} assumed that there exists an oracle agent that is capable of playing games perfectly and first used QA in TBGs by utilizing a network to determine the next action; the network was pretrained on the experience of the oracle agent.
They used knowledge graphs as state spaces for TBG agents and then showcased KG-A2C \cite{kga2c}, which was able to tackle a fully combinatorial action space for the first time.
To overcome the bottleneck problem and achieve improved sample efficiency, Ammanabrolu et al. \cite{QBERT} further proposed Q*BERT, which uses the pretrained ALBERT \cite{lan2019albert} language model for QA.
To correctly answer the given question, Q*BERT finetunes ALBERT on the SQuAD \cite{squad} and Jericho-QA datasets, which are specific to the TBG domain.
Given an observation, Q*BERT first produces a fixed set of questions that are sent to the fine-tuning ALBERT model, and the predicted answers are then used to update the knowledge graph as the current state for predicting the next action.
Compared to KG-A2C, Q*BERT reaches asymptotic performance faster, which indicates that the QA system leads to faster learning and thus improves the sample efficiency of the model.

Concurrently, Adhikari et al. \cite{gata} pointed out that prior works highly relied on heuristics that exploit games' inherent structures; for example, KG-A2C \cite{kga2c} uses hand-crafted heuristics to update its knowledge graph.
Thus, they tried to design more general and effective representations in an autonomous way by replacing heuristics with learning.
Adhikari et al. proposed the graph-aided transformer agent (GATA), which learns to construct and update graph-structured beliefs in an entirely data-driven manner.
The GATA combines the transformer version of LSTM-DQN with a dynamic belief graph, which is used as long-term memory for improving the action selection process by capturing the underlying game dynamics.
In particular, it consists of two modules: a graph updater and an action selector.
The graph updater is pretrained using two self-supervised training regimes.
The first method involves a decoder model that reconstructs text observations from the input belief graph by viewing this problem as a sequence-to-sequence task and using a transformer-based model:
\begin{equation}
    \mathcal{L}=- \sum_{i=1}^{L_{O_t}} \log p(O^i_t | O^1_t, \dots, O_t^{i-1}, \mathcal{G}_t, A_{t-1})
\end{equation}
where $\mathcal{G}_t$ is the belief graph, $A_{t-1}$ is the previous action, and $O_t$ is the observation.
The second method reformulates the first method as a contrastive prediction task and learns to maximize the mutual information between the predicted $\mathcal{G}_t$ and the text observations $O_t$.
The action selector encodes the belief graph $\mathcal{G}_t$ and text observation $O_t$ into hidden representations, and aggregates them using a bidirectional attention-based aggregator \cite{qanet}, which is trained via the double DQN \cite{DDQN} method.
The GATA achieves great performance and outperforms the baseline with ground-truth graphs, which evinces the effectiveness of generating graph-structured representations.

Thereafter, Tuli et al. \cite{LTL-GATA} thought that successful play also requires the ability to follow the instructions embedded within the given text.
Although instructions may not be directly bound to a reward, they can guide the actions of the agent to maximally achieve the ultimate goal.
However, the GATA \cite{gata} is not sensitive to the presence or absence of instructions, and it usually cannot complete tasks; thus, Tuli et al. proposed equipping the GATA with linear temporal logic (LTL) \cite{LTL}, a formal language that provides a mechanism for monitoring progress toward instruction completion.
Specifically, Tuli et al. first used LTL to augment the reward and define episode termination.
When the LTL instruction is satisfied, a bonus reward is given, and a penalty is given when the agent fails to complete an instruction.
If the LTL instruction is violated, the agent directly ends in a terminal state even if the environment has not indicated to do so.
Then, Tuli et al. augmented the original architecture to adapt to LTL instructions, including the LTL encodings of instructions and their in-game update process. %(shown in Figure \ref{fig:LTL-GATA}).
% \begin{figure}[!t]
%     \centering
%     \includegraphics[width=2.5in]{figs/LTL-GATA.jpg}
%     \caption{The framework of LTL-GATA (image from \cite{LTL-GATA}), which includes four encoders and one action selector.}
%     \label{fig:LTL-GATA}
% \end{figure}
%
Finally, the overall policy can be defined as $\pi(a | o, g, \phi)$, where $g, \phi$ are the belief state and LTL instruction, respectively.
Experiments showed that LTL augmentation results in large performance improvements and more effective instruction following and task completion.

Moving away from model-free methods, Liu et al. \cite{OOTD} focused on model-based planning and tried to learn a dynamics model from a TBG, which often includes partially observable states with noisy text dynamics.
Motivated by the object-oriented MDP (OO-MDP) \cite{oomdp} that factorizes world states into object states, Liu et al. designed the object-oriented text dynamics (OOTD) model, which includes graph representations for objects and independent transition layers, to predict belief states without knowing the rewards and observations during planning.
In particular, an OOTD model includes a transition model to predict the belief states of objects and a reward model that maps the sampled object states to the rewards, both of which are learned under the object-supervised and self-supervised settings.
The transition model first uses a graph decoder with the ComplEx scoring function \cite{ComplEx} to map the object states into a memory graph $h_{t-1}$ given the states of K objects $z_{t-1}=[z_{1,t-1},\dots, z_{K,t-1}]$ and then encodes $h_{t-1}$ into node representations $e_{t-1} = [e_{1,t-1},\dots, e_{K, t-1}]$ with a relational graph convolutional network (R-GCN) \cite{rgcn}.
Finally, to identify the affected objects given an action, OOTD uses a BIDAF network \cite{bidaf} and a group of transition layers to predict the belief states of objects with the ICM framework \cite{pearl2009causality}.
The reward model is formalized as follows:
\begin{equation}
    p_r(r_t | z_t, g_t) = \psi^r[z_t^{pool}, \psi^g(g_t)],
\end{equation}
where $g_t$ is the goal, $\psi^g$ is a transformer-based text embedding function and $\psi^r$ is implemented by a multilayer perceptron (NLP).
The empirical results obtained on the TextWorld benchmark \cite{textworld} demonstrated that OOTD-based planning methods are better than model-free baselines in terms of sample efficiency and performance.

\subsection{Navigation}
Learning to navigate in a photorealistic environment with visual-linguistic (VLN) clues has drawn increasing research interest \cite{R2R}, where agents need to understand language instructions, perceive their surrounding environments and perform navigation actions to reach the target goal. 
The VLN task can be formulated as a POMDP, where future observations are only dependent on the current state and action of the agent, and a visual observation only corresponds to a partial instruction, requiring the agent to memorize the navigation state and correctly localize the useful information for making the current decision.
To promote the development of the VLN field, several datasets and simulators have been developed, such as the natural language-guided R2R \cite{R2R} and Touchdown \cite{touchdown} navigation methods, dialog-based CVDN \cite{CVDN}, VNLA \cite{VNLA} and HANNA \cite{HANNA} navigation approaches, and the REVERIE \cite{reverie} navigation method for localizing remote objects.

The facts that each instruction only partially characterizes the trajectory and that visual states share various relationships with language instructions may be problematic for several methods \cite{misra2017mapping,wang2018look} that learn to understand the instructions from scratch.  
Hao et al. \cite{prevalent} proposed pretraining an encoder to align the representations of language instructions and visual states, which can alleviate the ambiguity of instructions; their approach is called PREVALENT.
Formally, the instructions are represented as a set of $\mathcal{X} = \{x_i \}_{i=1}^M$, each instruction contains $L_i$ word tokens $x_i = [x_{i,1}, \dots, x_{i, L_i}]$, and the agent learns to navigate through the policy $\pi$ given the training dataset $\mathcal{D}_E = \{\tau, x \}$ where $\tau$ is the expert trajectory:
\begin{equation}
    \max_{\theta} \mathcal{L}_{\theta} (\tau, x) = \log \pi_{\theta}(\tau | x) = \sum_{t=1}^T \log \pi_{\theta}(a_t | s_t, x),
\end{equation}
where $\theta$ is the policy parameter.
Hao et al. used two single-modal encoders with a cross-modal encoder, all of which are based on transformers, to construct a backbone and proposed two main tasks to pretrain the policy network: image-attended masked language modeling (MLM) and action prediction (AP).
MLM is similar to the pretraining process in BERT, which involves predicting the masked words based on the surrounding words by minimizing the negative log-likelihood:
\begin{equation}
    \mathcal{L}_{MLM} = -\mathbb{E}_{s \sim p(\tau), (\tau, x) \sim \mathcal{D}_E} \log p(x_i | x_{\backslash i}, s),
\end{equation}
where the recovery process is also based on the visual states $s$.
AP involves predicting an action without referring to the trajectory history by:
\begin{equation}
    \mathcal{L}_{AP} = -\mathbb{E}_{(a,s) \sim p(\tau), (\tau, x) \sim \mathcal{D}_E} \log p(a | x_{[CLS]}, s),
\end{equation}
where $[CLS]$ is the position whose output is used to predict the action.
The full pretraining objective is :
\begin{equation}
    \mathcal{L}_{\text{Pretraining}} = \mathcal{L}_{MLM} + \mathcal{L}_{AP}.
\end{equation}
%
%Comprehensive experiments demonstrate strong performance of PREVALENT in terms of adaptation speed and generalization. 
Other works have also emphasized the importance of pretraining; for example, PRESS \cite{press} understands language instructions through an off-the-shelf BERT model, VLN-BERT \cite{VLN-BERT} fine-tunes ViLBERT \cite{vilbert} on instruction-trajectory pairs to measure compatibility, and many methods \cite{chen2020uniter, li2020unicoder, su2019vl} pretrain a transformer on a large number of image-text pairs to learn generic cross-modal representations (these approaches are known as V\&L BERT).

Instead of using a pretrained model only for encoding language \cite{prevalent, press} or measuring compatibility \cite{VLN-BERT}, Hong et al. \cite{vlnObert} directly used a pretrained BERT model for learning to navigate with modifications to better adapt to VLN.
Hong et al. proposed the recurrent vision-and-language BERT model for navigation, which is able to effectively use the pretrained V\&L BERT model and no longer requires additional pretraining for the VLN task.
They augmented the original architecture with a recurrent function to memorize the historical representations and considered language tokens as keys and values but not queries in the self-attention modules to reduce the computational consumption \cite{lxmert}.
Experiments conducted on R2R \cite{R2R} and REVERIE \cite{reverie} showed that simple V\&L BERT plus a recurrent function is able to replace complex encoder-decoder models and achieve state-of-the-art results.

Furthermore, Chen et al. \cite{HAMT} intended to explicitly encode the historical information instead of using the recurrent states which might be suboptimal for capturing essential information for the subsequent trajectories \cite{fang2019scene}.
They proposed the history-aware multimodal transformer (HAMT), a fully transformer-based network for VLN tasks.
The HAMT uses a cross-modal transformer to capture the long-range dependencies of the current observation and instruction from the historical sequence and adopts a hierarchical version of the transformer to reduce the required computation expenses; this approach progressively learns the representations of single views, the spatial relationships among the views within a panorama, and the temporal relationships across different panoramas \cite{vivit}.
To learn better visual representations, the HAMT is also pretrained on several proxy tasks, such as AP, MLM, spatial relationship prediction (SPREL), instruction-trajectory matching (ITM), and fine-tuning for sequential action prediction with the RL+IL objective:
\begin{align}
    \theta \leftarrow \theta &+ \mu \frac{1}{T} \sum_{t=1}^T \nabla_{\theta} \log \pi(\hat{a}_t^h; \theta) (R_t - V_t) \notag \\
    &+ \lambda \mu \frac{1}{T^*} \sum_{t=1}^{T^*} \nabla_{\theta} \log \pi(a_t^*; \theta).
\end{align}
Through this method, the HAMT has strong performance in both seen and unseen environments in all tasks.
Similarly, Pashevich et al. \cite{E.T.} also proposed a multimodal transformer model (E.T.) to encode the full episode history and leverage synthetic instructions as the interface between the human and the agent to help the model learn more easily and generalize better.

\begin{figure}[!t]
    \centering
    \includegraphics[width=3.5in]{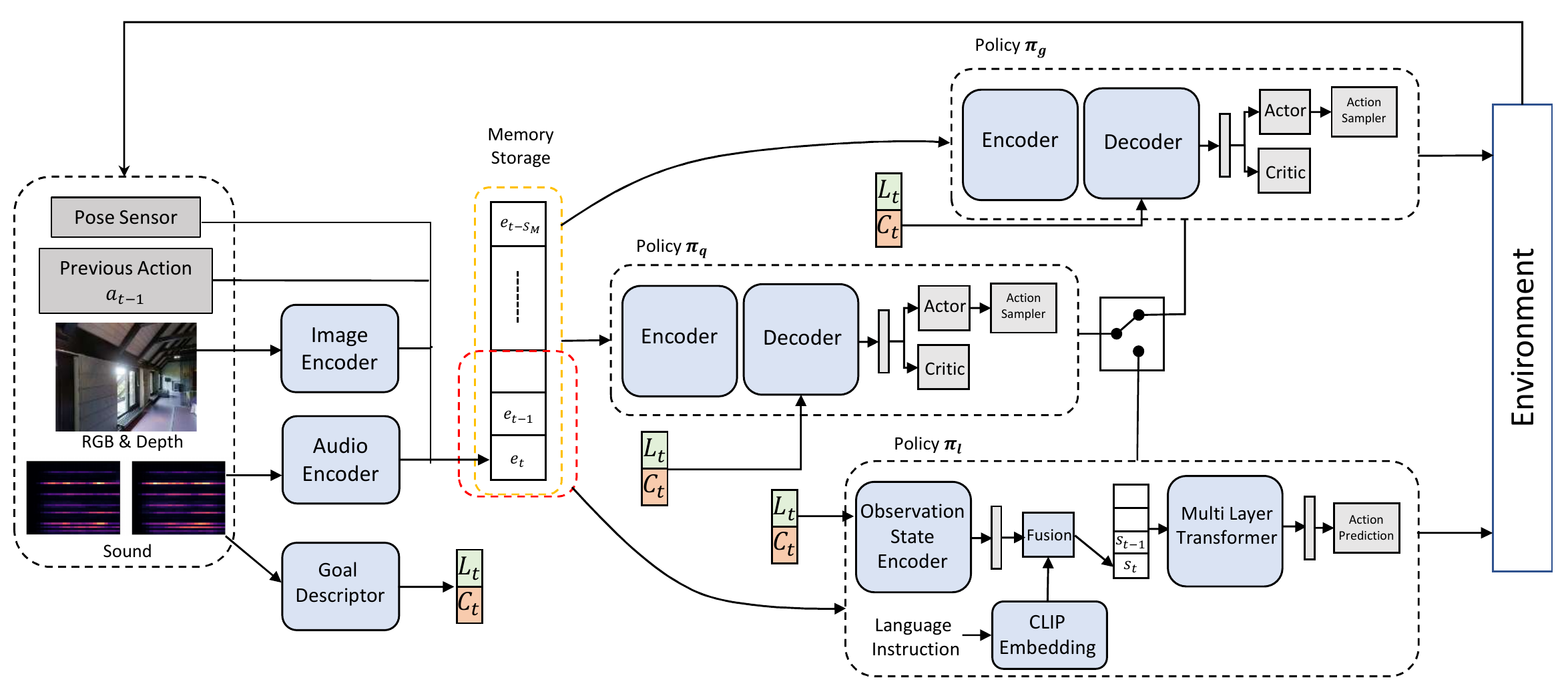}
    \caption{The framework of AVLEN (image from \cite{avlen}), which is a combination of query policy $\pi_q$, goal-based navigation policy $\pi_g$, and language-based navigation policy $\pi_l$}
    \label{fig:AVLEN}
\end{figure}

Afterward, Paul et al. \cite{avlen} considered a more complex and stochastic environment, the audio-vision-language environment, which is closer to the real-world navigation setting.
They presented AVLEN which guides the agent to localize an audio source in a realistic visual world.
The agent must not only learn audio-visual information \cite{chen2021semantic} but also seek help from an oracle for navigation instructions in the form of natural language.
It is important to decide when to query the oracle and when to follow audio-visual cues to reach the goal.
Paul et al. proposed a multimodal hierarchical RL framework, which consists of a high-level policy that selects which modality to use for navigation, and two low-level policies corresponding to different modalities based on the transformer architecture (shown in Figure \ref{fig:AVLEN}).
Experiments conducted on a Matterport 3D simulator \cite{R2R} enriched with audio events via the SoundSpaces framework \cite{soundspaces} showed that the audio-visual-language modeling agent leads to a clear performance improvement, especially when an oracle exists, which helps the agent deal with many challenging cases.

\subsection{Autonomous Driving}
The autonomous driving problem can be defined as point-to-point navigation in an urban setting while maintaining a safe distance from other dynamic agents and following traffic rules.
There are two main categories of learning approaches for self-driving vehicles (SDVs): supervised learning and RL.
From the development of the CARLA \cite{carla} and NoCrash \cite{NoCrash} driving simulators, a series of works have been proposed, such as SAM \cite{sam} that mimics driving methods, ST-P3 \cite{stp3} with explicitly intermediate representations for driving and NEAT \cite{neat} that adopts neural attention fields for reasoning scene structures.
However, these supervised learning methods require large human-labeled driving dataset and lack robust interpretability, and their performance is also limited by the offline handcrafted trajectory; hence, deep RL methods have attracted researchers' interest.
For example, Roach \cite{roach} used an RL-based model to provide a demonstration for an imitation learning agent, and Toromanoff et al. \cite{toromanoff2020end} presented a value-based Rainbow-IQN-Apex training method based on visual inputs.

An SDV usually takes both images acquired from cameras and the point cloud derived from the light detection and ranging (LiDAR) sensors as input.
A natural question is how to integrate representations from two or more modalities and maintain their respective advantages, which is called sensor fusion.
Prakash et al. \cite{transfuser} proposed the multimodal fusion transformer (TransFuser) model, which integrates representations from different modalities through a self-attention transformer and predicts future way points in an autoregressive manner based on a GRU model.
TransFuser is trained under imitation learning, where the policy involves learning to imitate the expert trajectory:
\begin{equation}
    \underset{\pi}{\text{argmin}}~~ \mathbb{E}_{(\mathcal{X}, \mathcal{W}) \sim \mathcal{D}} [\mathcal{L}(W, \pi(\mathcal{X}))],
\end{equation}
where $\mathcal{W} = \{(x_t, y_t) \}_{t=1}^T$ is the expert trajectory, $\mathcal{X}$ is the high-dimensional observation, and the loss function can be the $L_1$ loss, $L_2$ loss and so on.
However, Shao et al. \cite{InterFuser} thought that the TransFuser process harms sensor scalability and is limited to fusion between LiDAR and single-view images, which may result in difficulty when attempting to fully perceive the surrounding environment.
They further proposed an interpretable sensor fusion transformer (InterFuser), which is able to fuse information from multimodal multiview sensors for producing comprehensive feature representations.
Moreover, motivated by the information processing approach of humans\cite{radulescu2019holistic}, Shao et al. output an intermediate interpretable feature called a safety mind map, which provides information about surrounding dynamic agents and traffic signs and is exploited to constrain the actions within a safe action set by predefined criteria.
Extensive experiments conducted on CARLA showed that InterFuser achieves great performance in complex and adversarial urban scenarios.

%trajectory sequence prediction(SPLT related work)

For training datasets, algorithms need to learn behavioral priors from precollected data.
One option is offline RL, which learns a conservative policy for task-specific behaviors; however, it cannot solve problems where reward labels are not present.
Thus, behavior cloning is a more fitting option \cite{transfuser, InterFuser}.
However, these methods often make a fundamental assumption: the data only consist of unimodal expert trajectories, while the natural precollected data are often suboptimal and contain multiple modes of behavior.
To train models that can naively learn multimodal policy behaviors, Shafiullah et al. \cite{BeT} presented a behavior transformer (BeT), which is able to learn behaviors from distributionally multimodal data.
In particular, the BeT divides a continuous actions into two parts: a categorical variable denoting an "action center" by k-means \cite{macqueen1967classification} and a corresponding "residual action"; it then uses the transformer to map each observation to a categorical distribution over $k$ discrete action bins with the focal loss \cite{focalloss}:
\begin{equation}
    \mathcal{L}_{focal}(p_t) = -(1-p_t)^{\gamma} \log(p_t).
\end{equation}
An extra head is used to predict the offset with a loss akin to the masked multitask loss \cite{fastrcnn}:
\begin{equation}
    \text{MT-Loss} (\mathbf{a} , (\langle \hat{a}_i^{(j)} \rangle)_{j=1}^k) = \sum_{j=1}^k \mathbb{I} [ \lfloor \mathbf{a} \rfloor = j] \cdot || \langle \mathbf{a} \rangle - \langle \hat{a}^{(j)} \rangle ||_2^2,
\end{equation}
where $\mathbb{I}[]$ denotes the Iverson bracket, ensuring that the loss is only incurred from the ground-truth class of action $\mathbf{a}$.
Experiments conducted on CARLA showed that the BeT is able to cover all the modes of demonstration data.

\begin{figure}[!t]
    \centering
    \includegraphics[width=3.5in]{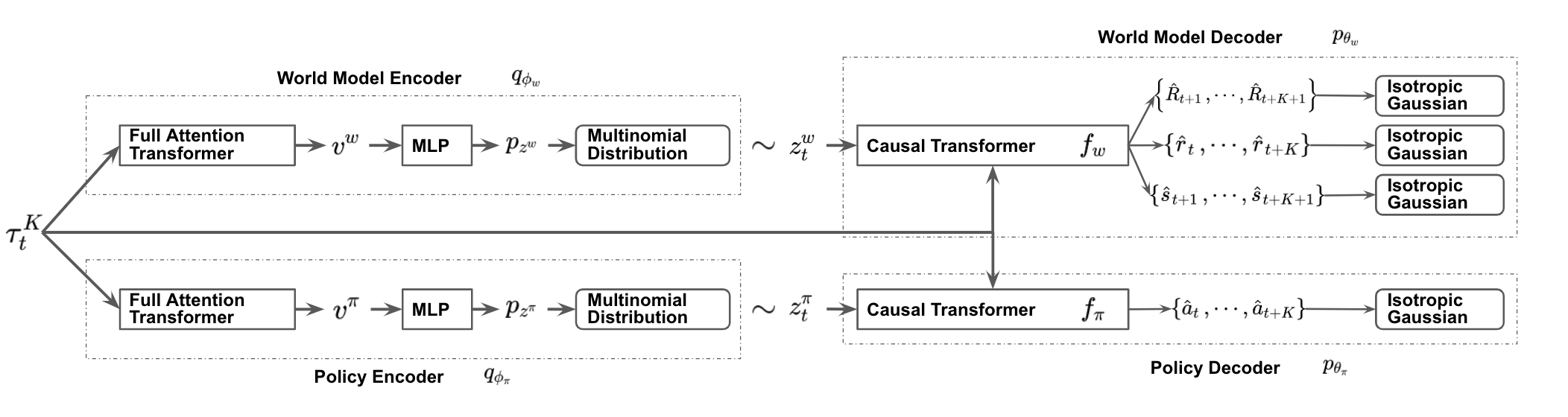}
    \caption{The framework of SPLT (image from \cite{SPLT}), which includes a world model to reconstruct the discounted returns, rewards, and states, and a policy model to reconstruct the action sequence.}
    \label{fig:SPLT}
\end{figure}

More recently, Villaflor et al. \cite{SPLT} tried to apply sequence modeling-based RL methods to stochastic safety-critical domains.
However, these methods jointly model the states and actions as a single sequencing problem which leads to overly optimistic behavior and may be dangerous in the autonomous driving domain.
Thus, Villaflor et al. developed a method called the separated latent trajectory (SPLT) transformer, to disentangle the effects of policies and world dynamics on the outcomes (shown in Figure \ref{fig:SPLT}).
Following the prior work \cite{tang2019multiple} in the autonomous driving domain, the SPLT transformer trains two separate transformer-based discrete latent variational autoencoders (VAEs) to represent the world and policy models, which are able to produce a range of candidate behaviors for the SDV and cover a majority of different potential responses from the environment.
By leveraging the two models, the SPLT transformer conducts robust planning at the test time with the intention of selecting a policy that will be robust to any realistic possible future in the environment.
Experiments conducted on CARLA demonstrated that the SPLT transformer is able to outperform the baseline approaches \cite{DT, TT, IQL} in a variety of autonomous driving tasks.

Many trajectory prediction works have also leveraged transformer neural networks \cite{ngiam2021scene, liu2021multimodal, quintanar2021predicting} to model long-range dependencies for prediction in complex traffic situations.
A transformer can also be used to model the relationships between an SDV and other dynamic agents and the static elements  (e.g., lanes, stop signs, and dynamic traffic light states) in the driving environment \cite{yuan2021agentformer, li2020end, park2020diverse}, which is beneficial for producing consistent predictions with respect to the surrounding traffic.
VAE models \cite{tang2019multiple, quintanar2021predicting} are also usually incorporated to cover the different possible modes of future traffic trajectories.

\subsection{Other Applications}
Automatic diagnosis is usually formulated as an MDP and solved by RL methods for policy learning \cite{hou2021imperfect, teixeira2021interplay}; however, these methods try to learn the symptom inquiry and disease diagnosis policy with the maximum reward rather than directly determining the doctor's diagnostic logic, and there are no optimal reward functions, which make it difficult to learn the optimal policy.
Considering that the diagnosis process can be seen as a sequence generation problem that includes symptoms and diagnoses, Chen et al. \cite{Diaformer} formulated automatic diagnosis as a sequence modeling problem, leveraging the transformer architecture (Diaformer) and training it in an autoregressive manner.
They also put forward an attention mask mechanism for automatic diagnosis: each explicit symptom can only see the other explicit symptoms, while each implicit symptom can see all the explicit symptoms and the previously generated implicit symptoms in an autoregressive manner.
To mitigate the discrepancy between the learned symptom sequence order and the disorder of realistic symptom inquiries, Chen et al. further proposed three orderless training mechanisms: sequence shuffling, synchronous learning, and sequence repetition.
Extensive experiments conducted on the MuZhi \cite{MuZhi}, Dxy \cite{Dxy} and Synthetic datasets \cite{Synthetic} demonstrated the efficacy of the Diaformer model.

The maintenance problem involves making maintenance decisions that minimize costs and keep the operators safe.
Several approaches provide failure predictions and remaining useful life (RUL) estimations for humans \cite{zheng2017long, wang2019remaining}; however, many operators believe that it is not sufficient to make maintenance decisions while knowing the likelihood of failure in the future and the RUL.
Motivated by the success of the DT\cite{DT} and TT\cite{TT}, Khorasgani et al. \cite{khorasgani2021offline} tried to solve maintenance decision making problems through offline RL methods to directly generate optimal maintenance actions such as "continuation" or "repair".
Given historical observations, actions, and RTGs, Khorasgani et al. learned to predict the action at each state.
Reliable RUL estimation models are commonly developed in industries.
When an RUL estimation is available, it can also be input into a decision model to achieve improved accuracy and simplify the training process, and this step is formulated as:
\begin{equation}
    \hat{a}_k = \texttt{model}(O_k^T, A_{k-1}^T, R_k^H, RUL_k),
\end{equation}
where $T$ is the length of the historical window and $H$ is the future horizon.
Experiments conducted on the C-MAPSS datasets \cite{C-MAPSS} showed that using offline RL to directly generate the optimal maintenance actions is competitive with using RUL estimation.

%% file: texfile/6.Discussion.tex
\section{Extension and Discussion}
\label{sec:ExtensionDiscussion}

%Transformer has made considerable progress in both the NLP field and the CV domain, so can the RL domain also use its powerful capabilities? 
%
%In order to explore and utilize the power of transformers, as summarized in this survey, many algorithms have been proposed in recent years. 
%
%From the beginning as an upgraded version of LSTM to remember historical information, to later used to model environmental information (e.g. dynamics transition, reward function). 
%
%The proposal of the sequence modeling perspective further introduces the ability of the transformer into the RL field, which has been fully developed in the NLP field. 
%
Transformers are becoming hot topics in the field of RL due to their strong performance and tremendous potential, as summarized in this survey.
A natural follow-up development direction concerns how to apply transformer-based methods to a wider range of RL systems, such as multiagent RL systems, which has been studied in several works.
%and currently there are some related research.
%
Nevertheless, the potential of transformers for performing RL has not yet been fully explored, meaning that several challenges still need to be resolved and several future prospects need to be considered.
In this section, we first introduce the transformer to the MARL extension, then discuss several challenges and finally provide insights on the future prospects of this field.

\subsection{Extension}
\begin{figure*}[!t]
\centering
\subfloat[]{\includegraphics[width=3.5in]{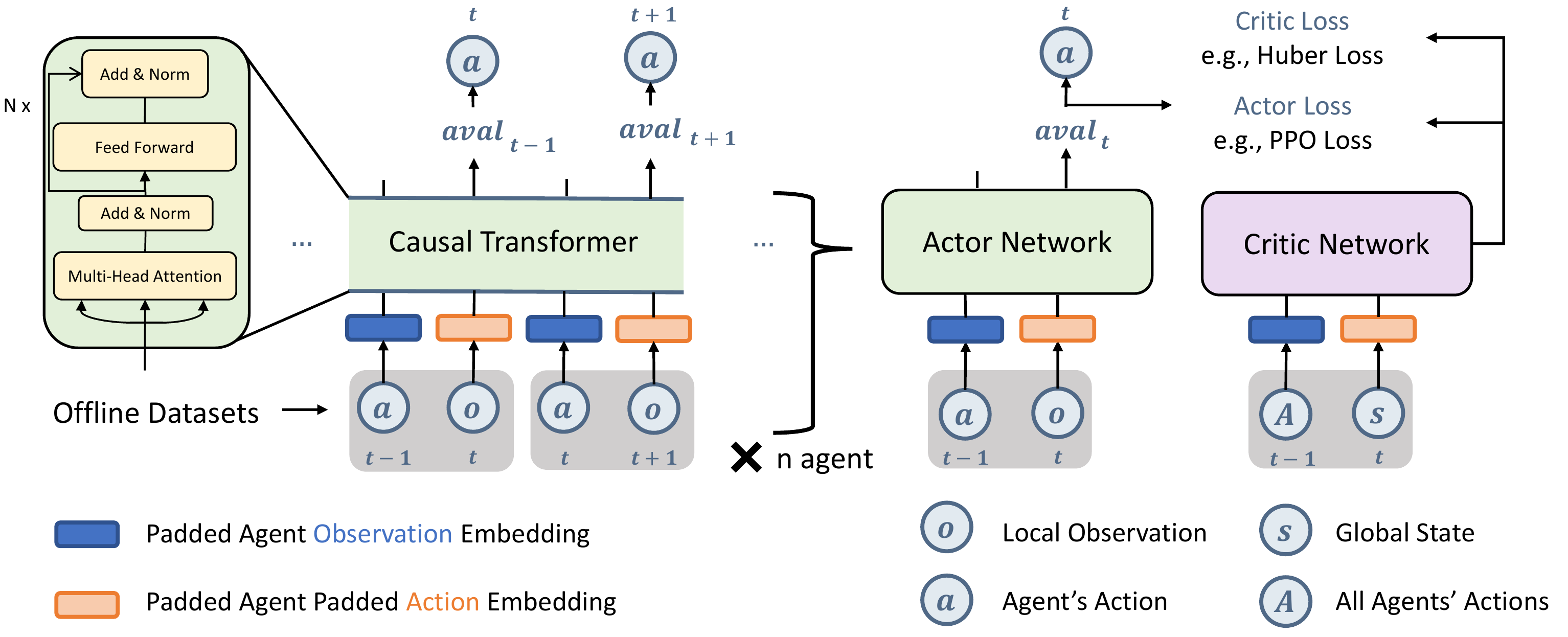}%
\label{fig:MADT}}
\hfil
\subfloat[]{\includegraphics[width=3.5in]{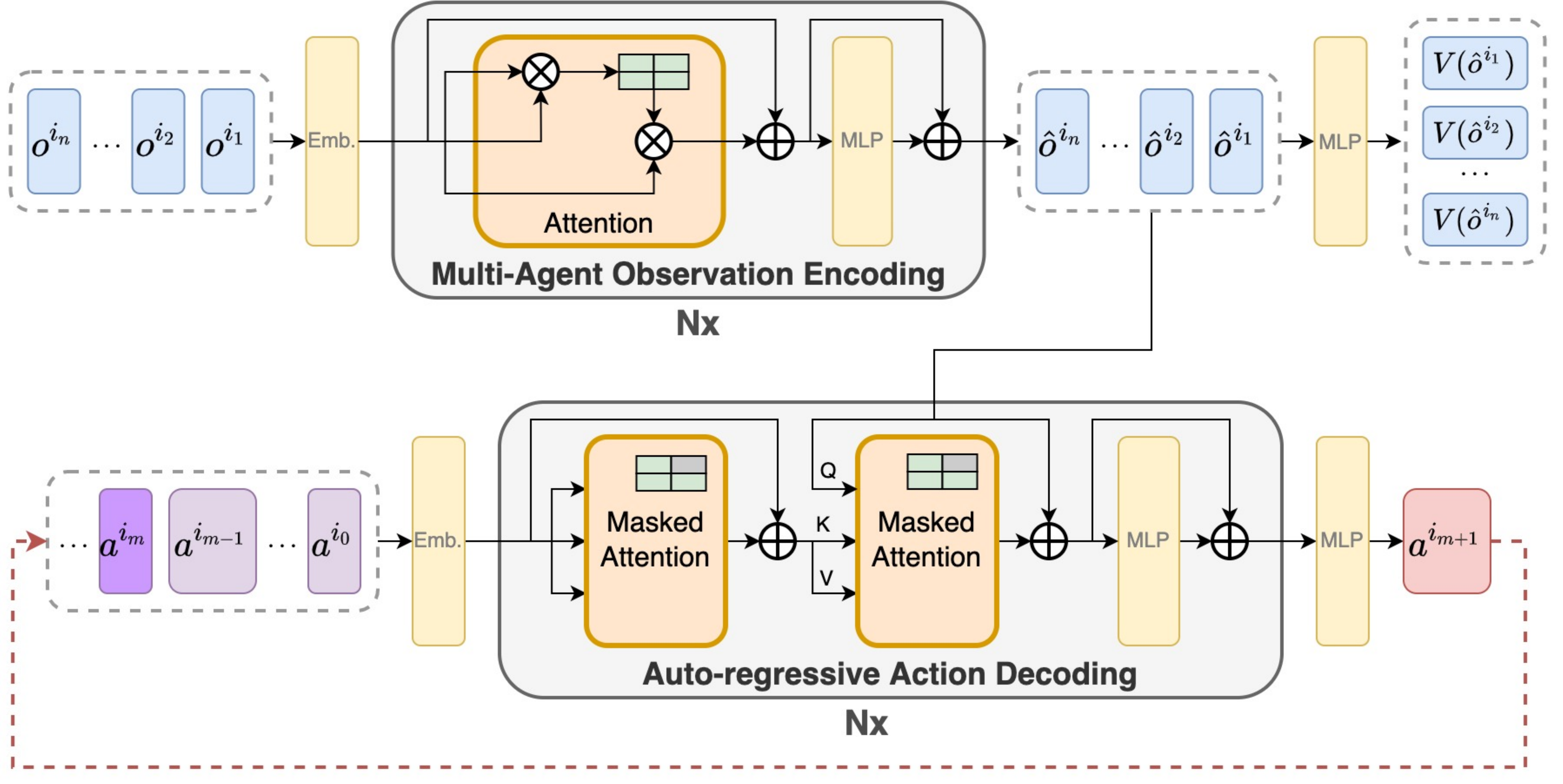}%
\label{fig:MAT}}
\caption{(a) The detailed structure for MADT (image from \cite{MADT}) which includes offline pretraining and online finetuning.~ (b) The encoder-decoder architecture for MAT (image from \cite{MAT}) which inputs a sequence of agents' observations and generates each agent's optimal action in an auto-regressive manner.}
\label{fig_sim}
\end{figure*}

The methods discussed above are all based on single-agent RL, and a natural extension is how to use transformers to solve MARL \cite{multi-agent-overview} in the form of sequence modeling.
The first work was conducted by Meng et al. \cite{MADT}, who studied the paradigm of offline pretraining with online fine-tuning in the MARL domain and provided large-scale datasets that were collected from running MAPPO \cite{MAPPO} on the well-known SMAC task \cite{SMAC}.
Treating the decision making process as a sequence modeling problem activates a novel pathway toward training RL systems on diverse datasets, which is important for the MARL problem since online exploration may not be feasible in many settings \cite{sanjaya2022measuring}.
Then, Meng et al. proposed a multiagent DT (MADT) which casts MARL as a conditional sequence modeling problem through an autoregressive transformer architecture (shown in Figure \ref{fig:MADT}).
Since MARL includes the local state for each agent, the author formulated the trajectory as follows:
\begin{equation}
    \tau^i = (x_1, \dots, x_t, \dots, x_T) ~~~ \text{where} ~ x_t = (s_t, o_t^i, a_t^i),
\end{equation}
where $s_t$ denotes the global state, $o_t^i$ denotes the local observation for agent $i$ at time step $t$, and $a_t^i$ is defined similarly. 
They employed the cross-entropy loss as the offline pretraining loss:
\begin{equation}
    L_{CE}(\theta) = \frac{1}{T} \sum_{t=1}^T P(a_t) \log P(\hat{a}_t | \tau_t , \hat{a}_{<t}; \theta),
\end{equation}
where $a_t$ is the ground-truth action, $\tau_t$ includes $\{s_{1:t}, o_{1:t} \}$ and $\hat{a}$ denotes the output of the MADT.
After conducting pretraining on an offline dataset, the transformer attains poor performance when directly using the above training method in an online environment since the agent still tries to select the actions that are similar to those in the offline dataset even if they are bad actions.
Thus, Meng et al. treated the pretrained transformer as the backbone of the actor and critic networks in their PPO algorithm \cite{PPO}.
Experimental results obtained on SMAC tasks showed that the pretrained MADT enjoys high sample efficiency and provides generalizability enhancements.

Afterward, Wen et al. \cite{MAT} proposed a multiagent transformer (MAT), which casts cooperative MARL as a sequence modeling problem from another perspective through an encoder-decoder transformer architecture (shown in Figure \ref{fig:MAT}).
Since the agents' interactions are important in multiagent systems, it is not guaranteed that the joint performance would improve when endowing each agent with a transformer policy and training them independently \cite{kuba2021trust}.
Motivated by the multiagent advantage decomposition theorem \cite{kuba2021settling}, which transforms joint policy optimization into a sequential policy search process according to Eq.\ref{maadt}, the MAT naturally solves this problem through sequence modeling and enjoys a monotonic performance improvement guarantee.
\begin{equation}
    \label{maadt}
    A_{\pi}^{i_{1:n}}(o, a^{i_{1:n}}) = \sum_{m=1}^n A_{\pi}^{i_m} (o, a^{i_{1:m-1}}, a^{i_m})
\end{equation}
Moreover, unlike the DT, which adopts offline imitation learning, the MAT is trained via online trial and error through a PPO-likely objective.
Extensive experiments conducted on SMAC, Multi-Agent MuJoCo\cite{de2020deep}, and Google Research Football \cite{kurach2020google} demonstrated that the MAT achieves superior performance and high data efficiency and is also an excellent few-shot learner.

To tackle the partial observability issue, gated RNNs are used to represent memory information in the single-agent \cite{hausknecht2015deep, kapturowski2018recurrent} and MARL settings \cite{omidshafiei2017deep, SMAC}.
Since transformers achieve more powerful performance than RNNs, Yang et al. \cite{ATM} proposed an agent transformer memory (ATM) network that incorporates transformer-based working memory \cite{WMG, dcrac} into MARL and utilizes an entity-bound action layer to incorporate semantic inductive bias into the action space.
In particular, the ATM first provides each agent with a memory buffer to store past information and updates the buffer through a first-in-first-out queue.
Then, the ATM gives each action its own semantic meaning to interact with different entities and uses the corresponding agents' embeddings to compute the associated Q-values or action probabilities:
\begin{equation}
    Q = Q(h_a, a),
\end{equation}
where $h_a$ is the embedding feature of the agent correlated to action $a$.
In experiments, ATM was used as the agent network and plugged into the QMIX \cite{qmix} and QPLEX \cite{qplex} algorithms, which are the most representative algorithms for SMAC, and the results demonstrated ATM's ability to increse its speed and achieve improved performance.
However, one of the limitations is that ATM needs to map actions to different agents manually which may be difficult or impossible in some environments.

Furthermore, Hu et al. \cite{updet} developed a model called the universal policy decoupling transformer (UPDeT), which is capable of handling multiple tasks at once.
The UPDet is based on the centralized training with decentralized execution (CTDE) architecture and imposes no limitation on the input or output dimensionality, which makes zero-shot transfer possible.
Each agent's value function is based on the transformer architecture and is mapped to a global Q-function through a credit assignment function $F$:
\begin{equation}
    Q_{\pi}(s_t, u_t) = F(q_1^t , \dots, q_n^t),
\end{equation}
where $u_t$ denotes the candidate actions and $q_i^t$ is the individual Q-function for agent $i$ at time step $t$.
The UPDet further matches the individual entities in observations to action groups and learns the relationships between the matched entity and other observation entities, which is called policy decoupling and is similar to the entity-bound action layer \cite{ATM}.
The final loss for optimizing the entire framework is the standard squared TD error for a DQN \cite{DQN}:
\begin{equation}
    L(\theta) = \sum_{i=1}^b [(y_i^{DQN} - Q(s, u; \theta))^2],
\end{equation}
where $b$ represents the batch size and $y_i^{DQN}$ denotes the target in the DQN algorithm.
By combining the transformer architecture and the policy decoupling strategy, the UPDet can be plugged into any existing algorithm (e.g., VDN \cite{vdn}, QMIX \cite{qmix} and QTRAN \cite{qtran}).
The experimental results obtained on SMAC \cite{SMAC} demonstrated the transfer capability of the UPDet in terms of both performance and speed.

\subsection{Challenges}
Although sequence modeling methods such as the DT \cite{DT} have achieved comparable performance on various tasks, these works are only the first steps in the RL domain and still have much room for improvement.

As shown in \cite{esper, rcsl}, these methods have high failure probabilities in stochastic environments, and many preconditions need to be satisfied to reach the optimal policy, such as deterministic dynamics and a known target return, where the target return corresponds to the distribution of the given dataset. 
Although Paster et al. \cite{esper} tried to condition the expected return to mitigate the negative effect of the uncertainty in the training dataset derived from a stochastic environment, their results were still inferior to those of a traditional RL algorithm, and it is less appropriate to condition the model on the mean return when the input to the agent is a visual goal.
There is still a long way to go to overcome these preconditions by changing the training method or improving the transformer structure.

The transformer architectures used in most works still follow the standard transformer for NLP \cite{vaswani2017attention}, and a transformer structure designed specifically for RL remains to be explored. 
Several works have tried to improve the architecture so that it can better adapt to RL. 
For example, Parisotto et al. \cite{GTrXL} modified the standard transformer architecture to obtain a more stable and faster training process, Shang et al. \cite{starformer} tried to incorporate a Markovian-like inductive bias into a transformer, and  Villaflor et al. \cite{SPLT} disentangled the policy and world dynamics generation processes to produce safer outcomes.
These improvements are attained in a task-driven manner rather than by designing a general transformer architecture with high RL performance; thus, the creation of an RL transformer is still an open problem waiting to be solved.

Finally, developing efficient transformer models for RL remains an open problem.
The network used in a deep RL algorithm usually contains only a few MLP layers, so it has short training and testing times, but the transformer is usually large and computationally expensive. 
For example, the transformer for the generalization task\cite{multigame_DT, Gato} usually needs to train on large-scale data, and its structure needs to correspondingly be very large to cover all possible situations. 
Although some methods try to reduce the size of their network structures through compression or distillation \cite{rae2019compressive, ALD}, they are still computationally expensive.
Consequently, efficient transformer models are urgently needed so that TRL methods can match the speed of traditional RL algorithms and can be deployed on resource-limited devices.

\subsection{Future Prospects}
To fully apply the capabilities of transformers to RL, we provide several potential directions for further study.

Treating RL as a sequence modeling problem simplifies many limitations encountered by traditional RL algorithms but also causes them to lose their advantages.
How to add the advantages of traditional RL algorithms to sequence modeling with a transformer is a problem worth studying. 
At the same time, it is noted that the original DT method \cite{DT} is based on RTGs for modeling future sequences. 
The introduction of RTG conditions makes the DT unable to learn the optimal results in many environments. 
How to correctly set these conditions and what function to use are both promising directions for future research. 
%
%Finally, it is noticed that many generalization models need to pretrain on large-scale data sets, and due to the shortage of large-scale datasets in the domain of RL, researchers usually train on language datasets \cite{chibiT, LID}, so it remains unexplored whether pretraining on different modality training sets will have an impact on downstream tasks.
%
Finally, how to conduct a pretraining process under the condition that the training dataset does not fully contain expert data is also a problem worth studying.

For traditional RL algorithms, the first problem that needs to be solved is the computational consumption of transformers.
Second, how to better use the transformer structure to adapt to a memory module is worth studying.
Moreover, in addition to serving as a memory module (replacing the LSTM) under POMDP conditions, what other roles the transformer can take on should also be further studied. 
At the same time, it is mentioned in \cite{AD} that when the given training dataset contains learning data and is sufficiently large, all RL algorithms can be imitated by the transformer structure.
Then, it is possible to design a general model that automatically switches the corresponding traditional RL algorithm according to the given tasks, but this has has not yet been studied.

For real-world applications, in addition to using the transformer structure as a multimodal multitask feature extraction method, it is also worth studying how to apply the transformer structure to decision networks.
It is necessary to fully understand the causes of errors and the possible negative consequences generated by transformers, which are critical to the security of practical applications.
%
%Although the task dependence of traditional RL algorithms makes it impossible to become a general model (e.g. in the domain of automatic driving), the introduction of the transformer structure makes it possible.
%

Note that the experimental environments used by the different methods presented in this survey are basically different, so it is difficult to obtain a unified benchmark for intuitively comparing different methods.
Here, we recommend that follow-up research adopt a unified baseline and metric.
Architecture enhancement methods are usually evaluated in environments where the critical observations often span the entire episode; thus, the DMLab-30 environment \cite{beattie2016deepmind} is a better choice in cases where memory information is important for the agent to make optimal decisions.
For trajectory optimization, since most methods are based on offline RL, the D4RL dataset \cite{d4rl} is currently the most commonly used environment and has a large number of offline trajectories.
To verify the generalization ability of an algorithm, it is necessary to train a single agent for use in different environments, where the Atari games suite and gym\_minigrid environment \cite{chevalier2018minimalistic} are currently fully utilized since they both contain many different tasks.
Different applications have their own representative test environments, which are not be further introduced here.
In addition to the content introduced above, TRL still has great potential in other fields and deserves to be fully studied.

\section{Conclusion}
In this survey article, we provide a comprehensive overview of transformers' involvement in RL and the associated realistic applications.
First, we provide the preliminaries of RL, especially offline RL, and the architecture of the transformer. 
Then, we review existing works concerning transformers in architecture enhancement and trajectory optimization methods and give a brief overview of TRL in the robotic manipulation, TBG, navigation, and autonomous driving domains.
Finally, we share our perspective on the open problems in the field, including possible domains for extension, several potential challenges, and promising future research directions.